\documentclass[sigconf,nonacm]{acmart}

\renewcommand\footnotetextcopyrightpermission[1]{}

\AtBeginDocument{%
  }

\usepackage{svg}
\usepackage{multirow}
\usepackage{colortbl}
\usepackage[table,dvipsnames]{xcolor}
\definecolor{lightgreen}{RGB}{220, 255, 220}
\usepackage{siunitx}
\usepackage{lipsum}
\usepackage{graphicx}
\usepackage{array}
\usepackage{booktabs}
\usepackage{float}
\usepackage{listings}
\usepackage{xspace}
\usepackage{enumitem}
\usepackage{longtable}
\usepackage[normalem]{ulem}
\usepackage{pifont} 
\usepackage{makecell}

\definecolor{dypink}{HTML}{ec008c}

\definecolor{mmblue}{HTML}{0077CC}

\newcommand{\parasum}[1]{\textbf{\textit{#1}}}

\begin{document}
\newcommand{\sys}{VTBench}
\newcommand{\syss}{VTBench\xspace}

\title{VTBench: A Multimodal Framework for Time-Series Classification with Chart-Based Representations}


\author{Madhumitha Venkatesan}
\affiliation{%
  \institution{University of California, Davis}
  \city{Davis}
  \country{United States}}
\email{mvenkat@ucdavis.edu}

\author{Xuyang Chen}
\affiliation{%
  \institution{University of California, Davis}
  \city{Davis}
  \country{United States}}
\email{lanche@ucdavis.edu}

\author{Dongyu Liu}
\affiliation{%
  \institution{University of California, Davis}
  \city{Davis}
  \country{United States}}
\email{dyuliu@ucdavis.edu}

\renewcommand{\shortauthors}{Venkatesan et al.}
\begin{abstract}
Time-series classification (TSC) has advanced significantly with deep learning, yet most models rely solely on raw numerical inputs, overlooking alternative representations. While texture-based encodings such as Gramian Angular Fields (GAF) and Recurrence Plots (RP) convert time series into 2D images, they often require heavy preprocessing and yield less intuitive representations. In contrast, chart-based visualizations offer more interpretable alternatives and show promise in specific domains; however, their effectiveness remains underexplored, with limited systematic evaluation across chart types, visual encoding choices, and datasets. In this work, we introduce \textit{\sys}, a systematic and extensible framework that re-examines TSC through multimodal fusion of raw sequences and chart-based visualizations. \syss generates lightweight, human-interpretable plots--line, area, bar, and scatter, providing complementary views of the same signal. We develop a modular architecture supporting multiple fusion strategies, including single-chart visual–numerical fusion, multi-chart visual fusion, and full multimodal fusion with raw inputs. Through experiments across 31 UCR datasets, we show that: (1) chart-only models are competitive in selected settings, particularly on smaller datasets; (2) combining multiple chart types can improve accuracy by capturing complementary visual cues; and (3) multimodal models improve or maintain performance when visual features provide non-redundant information, but may degrade accuracy when they introduce redundancy. We further distill practical guidelines for selecting chart types, fusion strategies, and configurations. \syss establishes a unified foundation for interpretable and effective multimodal time-series classification.

\end{abstract}

 


\keywords{Time-Series Classification, Deep Learning, Multimodal Learning, Visual Representations, Feature Fusion, Benchmarking}



\maketitle
\section{Introduction}
\label{intro}

Time-series classification (TSC) is central to many high-impact applications, including healthcare~\cite{pyakillya2017deep}, finance~\cite{xianya2019stock}, environmental monitoring~\cite{gomez2016optical}, human activity recognition~\cite{ding2020rfnet}, and industrial diagnostics~\cite{Alnegheimish2022SintelAM}. Deep learning methods such as LSTMs~\cite{karim2017lstm}, Temporal Convolutional Networks (TCNs)~\cite{lea2017temporal}, and Transformers~\cite{wang2024medformer} have demonstrated strong performance on TSC tasks by modeling temporal dependencies in raw sequences. 
More recently, time-series foundation models~\cite{liang2024foundation,ma2024surveypretrained,kottapalli2025foundation} have leveraged large-scale pretraining and often match---or occasionally surpass---the performance of conventional models.

In parallel, an emerging research direction has explored transforming time-series data into image representations. Early work focused on texture-based encodings such as Gramian Angular Fields (GAF)~\cite{wang2015imaging}, Recurrence Plots (RP)~\cite{Zhang2020EncodingTS}, and Markov Transition Fields (MTF)~\cite{hatami2018classification}, which are then processed using image classification models. More recent efforts have shifted toward chart-based visualizations--particularly line plots~\cite{li2023time,rodrigues2021plotting,zhao2025images}--which provide more intuitive and human-aligned representations. These formats not only enhance interpretability but also enable the reuse of powerful vision backbones. For example, ViTST~\cite{li2023time} converts multivariate time series into grid-aligned line charts and applies a Vision Transformer, achieving strong performance on irregularly sampled, large-scale healthcare and activity datasets. Despite these advances, existing studies remain narrow in scope: they typically focus on a single chart type, evaluate only on large datasets, and treat visual inputs as direct replacements rather than complements to raw numerical sequences.

Yet, several important research questions remain open: \textit{Which chart types are most effective under different conditions? How does their utility vary across domains and data characteristics? Can chart-based visualizations meaningfully complement--rather than replace--raw numerical inputs? How do factors such as training set size and sequence length impact their effectiveness? What unique advantages do visual chart representations offer over traditional approaches?} Addressing these questions requires a systematic evaluation of chart types across diverse datasets, scales, and modeling scenarios.

We address these gaps by introducing \sys, a systematic benchmark for chart-based and multimodal time-series classification. \syss evaluates four standard chart types—line, bar, scatter, and area—across 31 datasets from the UCR archive~\cite{dau2019ucr}, covering a wide range of domains, sequence lengths, and dataset scales. We investigate these visual representations both individually and in combination with raw time-series signals. Our architecture pairs a CNN for chart images with a lightweight encoder for numerical sequences, integrating both through a modular fusion mechanism. \syss supports multiple configurations, including single-chart models, multi-view chart fusion, and full multimodal fusion with raw inputs. 
Our contributions are twofold:
\begin{itemize}[leftmargin=10pt]
\item \textbf{\syss framework.} We present the first systematic benchmark for chart-based and multimodal time-series classification. \syss supports four interpretable chart types (line, bar, scatter, area), integrates visual and numerical modalities via lightweight fusion, and enables configurable experimentation across single-chart, multi-chart, and multimodal settings. All code, chart pipelines, and training protocols are released for reproducibility and extensibility\footnote{\url{https://github.com/via-cs/vtbench}}



\item \textbf{Empirical findings.} We conduct large-scale experiments on 31 UCR datasets and observe that: (i) chart-only models can achieve competitive performance in selected settings, particularly on smaller datasets; (ii) multi-chart fusion can provide dataset-dependent gains by leveraging complementary information across chart types; and (iii) multimodal fusion--combining one or more charts with raw inputs--may improve, maintain, or degrade performance depending on whether the visual representations contribute non-redundant information. These findings offer a systematic perspective on when and how chart-based representations are effective, addressing key open questions in chart selection, visual encoding design, and fusion strategies, and providing practical guidance across diverse datasets and \mbox{application domains}. 
\end{itemize}

\section{Related Work}
\label{related_work}

\textbf{Numerical Data-Based Approaches.}
Several recent surveys have extensively evaluated time-series classification (TSC) methods~\cite{middlehurst2024bake, irani2025time, ma2024surveypretrained}. Middlehurst et al.~\cite{middlehurst2024bake}, for example, revisited the ``bake off'' comparison using an expanded UCR archive~\cite{dau2019ucr} (112 univariate datasets) and confirmed that hybrid ensembles—combining diverse feature-based classifiers—remain the most effective. Among them, HIVE-COTE 2.0 (HC2) \cite{middlehurst2021hive} currently achieves the best average accuracy across both univariate (UCR) and multivariate (UEA) benchmarks~\cite{dau2019ucr}. Other competitive models include deep neural networks such as Omni-Scale CNN~\cite{tang2020omni}, MultiROCKET+Hydra~\cite{Dempster2022HydraCC}, and InceptionTime~\cite{IsmailFawaz2019InceptionTimeFA}, which offer strong performance with improved efficiency. 
More recently, general-purpose time-series representation learning models have emerged, including TimesNet~\cite{TimesNet} and iTransformer~\cite{Liu2023iTransformerIT}, which aim to serve as adaptable backbones across tasks. Time-series foundation models~\cite{kottapalli2025foundation,ma2024surveypretrained,liang2024foundation} extend this line by leveraging pre-trained language or vision architectures for downstream applications. While these generic approaches show promise—particularly on complex \textit{multivariate} datasets—they have yet to surpass HC2 and other strong baselines (e.g., SoftShape~\cite{Liu2025LearningSS}) on broad \textit{univariate} benchmarks.


In our experiments, we benchmark a set of representative TSC algorithms—including HC2 and efficient deep models—on UCR datasets spanning small, medium, and large scales. This provides a competitive and practical baseline for evaluating our chart-based multimodal approach, without relying on large foundation models that require substantial resources.

\textbf{Visual Image-Based Approaches}
This line of research transforms time-series data into images, leveraging advances in computer vision for TSC. Classic methods encode temporal dynamics as 2D textures using transformation functions such as GAF, RP, or MTF~\cite{wang2015encoding, hatami2018classification, Zhang2020EncodingTS}, allowing image classifiers to extract discriminative patterns. These representations can highlight periodic or recurrent structures not easily captured in raw sequences. However, they often involve substantial preprocessing and hyperparameter tuning, which limits their use in real-time or resource-constrained settings.
More recent studies have explored visual chart representations applied to financial~\cite{chen2020encoding, moghaddam2021image} and healthcare~\cite{jun2018ecg} data. Rodrigues et al.~\cite{rodrigues2021plotting} demonstrated that simple line charts combined with CNNs can achieve competitive results. Extending this idea, Li et al.~\cite{li2023time} introduced ViTST, which converts multivariate time series into line-chart images and applies a Vision Transformer to capture global structure. While effective, such transformer-based models require large datasets and significant compute.

Notably, prior work has not systematically evaluated \textit{chart-based} representations across different chart types, encoding settings, and dataset characteristics. In contrast, we explore a range of charting techniques and design choices to assess their influence on classification. We adopt a standard CNN as the image encoder, providing a practical trade-off between efficiency and capacity for UCR-scale datasets, and enabling direct comparison with numerical baselines.

\textbf{Hybrid and Multimodal Models}
A growing area of research explores combining multiple representations of time series to improve classification and interpretability. Early fusion methods concatenate raw features with handcrafted descriptors at the input level \cite{liu2019salient, hoang2021deep} but often suffer from scale mismatch and weak modality interactions~\cite{dietz2024intermodal}. More effective are \textit{intermediate fusion} architectures, where each modality is encoded separately and merged later in the feature space. Recent multi-branch models follow this design, using separate encoders for different views and fusing embeddings via concatenation or attention~\cite{liu2024multi, liu2024efficient}. Extensions include Bayesian optimization for fusing visual features~\cite{mariani2025fusion} and meta-feature fusion for few-shot TSC~\cite{park2023meta}. Zhao et al.~\cite{zhao2025images} further apply large vision models to a range of image-based encodings—including line plots, heatmaps, and spectrograms—demonstrating their potential for time-series analysis. We adopt the intermediate-fusion paradigm in developing \syss, which integrates numeric and chart-based visual representations. To our knowledge, no prior work has systematically evaluated such a strategy across standard benchmarks.
\section{Methodology}
\subsection{Problem Formulation}
Let $\mathcal{X} = \{x_1, x_2, \ldots, x_n\}$ denote a collection of time-series instances, where each instance $x_i \in \mathbb{R}^{T \times d}$ represents a univariate time series of length $T$ (i.e., $d = 1$). Each instance is associated with a class label $y_i \in \mathcal{Y}$, where $\mathcal{Y} = \{1, \ldots, C\}$, and $C$ denotes the number of classes, accommodating both binary ($C=2$) and multi-class classification tasks. The objective of time-series classification (TSC) is to learn a function $f: \mathbb{R}^{T} \rightarrow \mathcal{Y}$ that maps input sequences to their corresponding class labels with high accuracy.

In this work, we reframe TSC as a visual recognition problem by transforming each time-series instance $x_i$ into one or more chart-based image representations, denoted by $\mathcal{I}(x_i)$. These representations include standard plot types, such as line, bar, area, and scatter charts. Let $\{\mathcal{I}_k(x_i)\}_{k=1}^K$ represent the set of $K$ distinct chart images generated from $x_i$; in our experiments, we use $K=4$, though this can be extended to additional views or styles. Each image $\mathcal{I}_k(x_i)$ is processed by a dedicated convolutional encoder $\phi_k(\cdot)$ to extract high-level visual features. These features are then fused and passed to a classifier $g(\cdot)$ to yield the final prediction:
\[
\hat{y}_i = g\left( \text{Fuse} \left( 
\phi_1(\mathcal{I}_1(x_i)),\ 
\phi_2(\mathcal{I}_2(x_i)),\
\ldots,\ 
\phi_K(\mathcal{I}_K(x_i)) 
\right) \right).
\]
We further extend this formulation by optionally incorporating the raw numerical sequence $x_i$ as an additional modality. This sequence is encoded via a separate numerical encoder $\phi_{K+1}(\cdot)$. The resulting multimodal features are fused at the feature level, allowing the model to leverage both visual and temporal cues for classification.

\subsection{Framework Design Rationales}

\begin{figure}[t]
  \centering
  \includegraphics[width=0.48\textwidth]{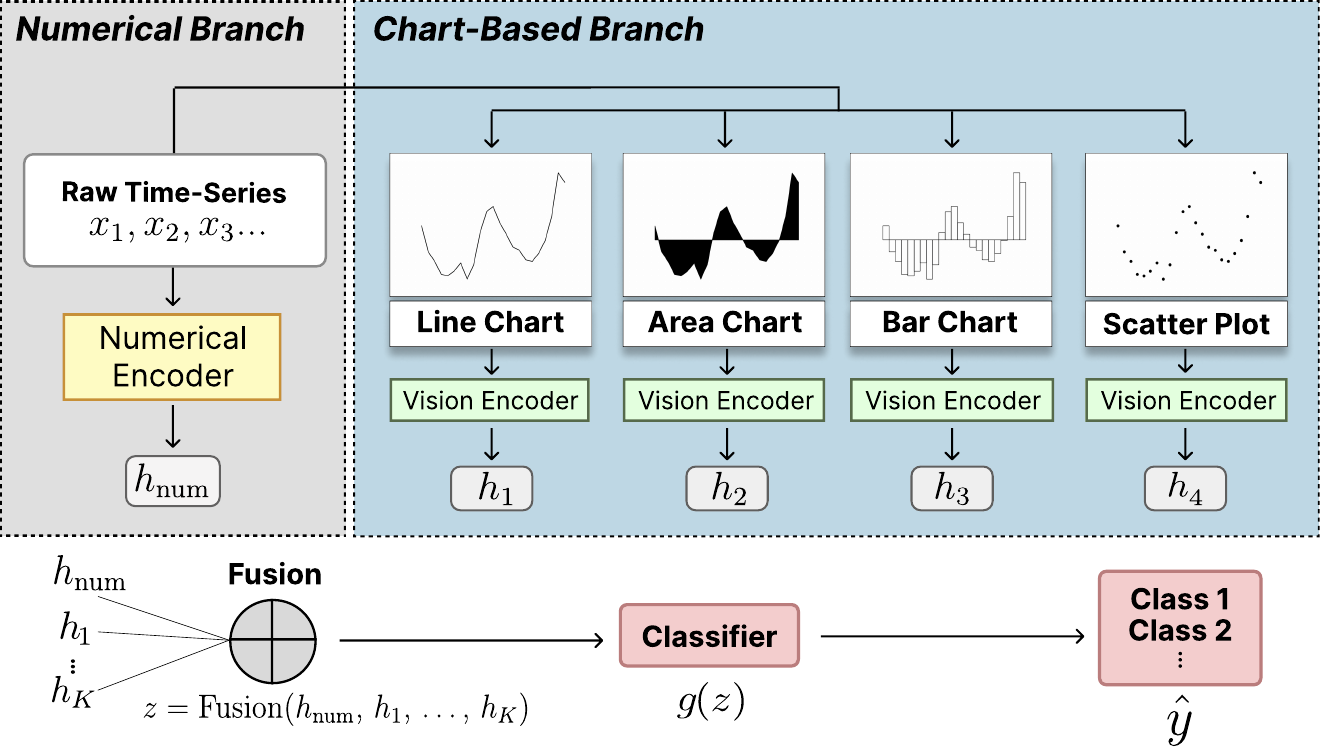}
    \caption{\syss converts a univariate time-series input into multiple chart-based visual representations, each processed by a vision encoder (e.g., CNN) to extract modality-specific features. The numerical signal is encoded using a temporal encoder (e.g., Transformer or OS-CNN~\cite{tang2020omni}) to capture temporal dependencies. The resulting features are fused and passed to a classifier for label prediction.}
  \label{fig:system}
  \vspace{-2em}
\end{figure}

We design \textbf{\sys} as a modular benchmarking framework to evaluate chart-based visual encodings for time-series classification (TSC)—both independently and fused with raw numerical inputs. Unlike prior work that treats visualizations as substitutes for raw signals or focuses on a single chart type, \syss supports flexible multimodal fusion. Figure~\ref{fig:system} illustrates the system architecture.

We adopt four commonly used chart types—\textit{line}, \textit{area}, \textit{bar}, and \textit{scatter}—chosen for their complementary visual semantics and prevalence in real-world time-series analysis~\cite{sarikaya2018we,xu2021mtseer,masry2022chartqa}. Each chart type emphasizes different signal characteristics: continuity (line, area), discrete changes (bar), and local outliers or sparsity (scatter), analogous to multiview learning in computer vision~\cite{li2018survey,bai2021correlative}. Given a univariate time series $x_i$, \syss generates four chart images using \texttt{matplotlib}, rendered at 128$\times$128 resolution with configurable settings for color (RGB or grayscale) and annotations (with or without axes). Each chart is passed to a dedicated CNN encoder (shallow or deep), extracting modality-specific features. These features are fused via either simple concatenation or dynamic weighted fusion (Section~\ref{sec:fusion}), enabling expressive multiview embeddings.

In multimodal configurations, \syss additionally encodes the raw sequence $x_i$ through a separate numerical branch—using FCNs, Transformer encoders, or state-of-the-art TSC models—to capture precise temporal dynamics. While CNNs emphasize chart-specific structural patterns (e.g., edge-like features in bar charts~\cite{zhou2018interpretable}), the numerical branch retains high-fidelity signal information. Combining these complementary views yields richer, more robust representations. \syss supports three architectural configurations: (1) \textbf{Single-modal}, using a single chart type; (2) \textbf{Multi-chart}, fusing all chart views; and (3) \textbf{Multimodal}, integrating both chart and raw inputs. This modular setup enables systematic evaluation of encoding and fusion strategies under consistent \mbox{experimental controls}.

Finally, \syss is designed for extensibility. New chart types, visual encoders (e.g., ResNet~\cite{he2016deep}, Vision Transformers~\cite{dosovitskiy2020image}), and additional modalities (e.g., frequency-domain features~\cite{alaa2021generative}, handcrafted features~\cite{middlehurst2021hive}, or metadata~\cite{xu2018raim}) can be easily incorporated. Its configurable architecture provides a general foundation for interpretable, multimodal time-series modeling.

\subsection{\syss Modular Architecture}
\label{sec:model_architecture}
\syss offers a flexible and modular architecture that supports both unimodal and multimodal input configurations. Each input modality—whether a chart-based image or a raw numerical time series—is processed through a dedicated feature extraction branch.

\paragraph{CNN Encoders for Chart-Based Inputs.}
Chart representations (line, bar, area, scatter) are rendered as $128 \times 128$ RGB images and processed by dedicated parallel CNN branches. CNNs are particularly well-suited for this task because they effectively capture spatially localized patterns—such as slopes, peaks, and clustered regions—that naturally emerge in time-series visualizations. The convolutional filters progressively extract higher-order features, enabling meaningful abstraction without handcrafted design. We evaluate two backbone variants with different capacity profiles:
\begin{itemize}[leftmargin=10pt]
    \item \textbf{ShallowCNN (3-layer):}
    A lightweight encoder comprising three convolutional blocks with filter sizes ${16, 32, 64}$, each using $3 \times 3$ kernels, stride 1, padding 1, followed by batch normalization, ReLU activation, and $2 \times 2$ max pooling. The resulting feature map is flattened and projected to a 64-dimensional embedding via a fully connected layer with ReLU activation and dropout ($p=0.5$). This configuration provides a low-parameter baseline ($\approx$286K params). to assess whether minimal spatial processing is sufficient for chart-based classification.
    
    \item \textbf{DeepCNN (5-layer VGG-style\cite{simonyan2014very}):}
    A deeper architecture with five convolutional blocks (filters sizes 16, 32, 64, 128, 256), kernel size $3 \times 3$, stride 1, padding 1), each followed by batch normalization, ReLU, and $2 \times 2$ max pooling. The output is flattened and passed through two fully connected layers (512 $\rightarrow$ 256 units) with ReLU activations and dropout ($p=0.5$) to yield a compact 256-dimensional embedding. This deeper encoder captures richer spatial hierarchies and complex features while remaining computationally lightweight ($\approx$1.2M params).
\end{itemize}

\paragraph{Numerical Feature Encoder.}
To facilitate joint modeling of visual and temporal modalities, \syss includes an optional numerical branch that processes the raw time-series input directly. We support the following encoder architectures:
\begin{itemize}[leftmargin=10pt]
    \item \textbf{Fully Connected Network (FCN):}
    A simple two-layer fully connected multilayer perceptron (MLP) with ReLU activations, used as a low-complexity baseline.
    
    \item \textbf{Transformer Encoder:} 
    A standard Transformer model \cite{vaswani2017attention} using self-attention to capture long-range \mbox{temporal dependencies}.
    
    \item \textbf{OS-CNN (SOTA Numerical Model):} 
  A state-of-the-art convolutional TSC model \cite{tang2020omni}, chosen for accuracy, efficiency, and architectural parity with our chart-based CNNs. Unlike heavy ensembles (e.g., InceptionTime) or feature-based methods (e.g., ROCKET), OS-CNN balances expressiveness and fairness, avoiding excess parameters or stochastic features that bias comparisons. It is robust on small-to-medium datasets, interpretable, and lightweight, making it a principled baseline for isolating representation effects versus chart-based encoders.
\end{itemize}
The output of this numerical branch is a vector embedding $h_{\text{num}}$, which is fused with visual features in multimodal settings.

\paragraph{Fusion Strategies.}
\label{sec:fusion}
We adopt two simple and widely used fusion mechanisms \cite{zhang2024multimodal, li2024multimodal, zeng2019deep} to combine modality-specific features. These strategies are deliberately simple to avoid the confounding effects of large fusion networks, keeping comparisons focused on representation quality.
Specifically, we evaluate:

\begin{itemize}[leftmargin=10pt]
    \item \textbf{Concatenation:} 
    A straightforward baseline that concatenates all modality-specific embeddings:
    \[
    z = [h_\text{num},h_1, \ldots, h_K]
    \]
    where $h_K$ is the output of the $k$-th branch, $h_\text{num}$ the optional numerical branch, and $z$ is the fused feature vector passed to the classifier. This design isolates the effect of added modalities without introducing extra parameters or weighting biases.

\item \textbf{Dynamic Weighted Fusion:} 
    To adaptively integrate heterogeneous modalities, we introduce a lightweight attention-based fusion mechanism that assigns learnable importance weights to each branch. Let $h_k \in \mathbb{R}^d$ denote the embedding from the $k$-th chart branch ($k=1\dots K$), and let $h_{\text{num}}$ represent the optional numerical branch. We compute attention weights as:
    \[
    \alpha_k = \frac{\exp(w_k^\top h_k)}{\sum_{j=1}^{K}\exp(w_j^\top h_j)\;+\;1_{\text{num}}\exp(w_{\text{num}}^\top h_{\text{num}})},
    \]
    where $w_k \in \mathbb{R}^d$ are trainable attention vectors, and $1_{\text{num}}\in\{0,1\}$ indicates the presence of a numerical branch. The fused representation is:
    \[
    z = \sum_{k=1}^{K}\alpha_k\,h_k \;+\; \alpha_{\text{num}}\,h_{\text{num}},
    \]
    where $\alpha_k \in [0,1]$ and $\sum_k \alpha_k = 1$, ensuring normalized contributions across modalities. We normalize contributions via a softmax, letting the network naturally emphasize informative modalities while maintaining non-negative, bounded weights. We intentionally avoid additional balancing regularizers, as they may artificially enforce uniform contributions even when one modality is genuinely more discriminative. This design choice aligns with our benchmark goal of isolating representation quality without injecting fusion-specific biases.
\end{itemize}
    
\noindent
We also avoid complex fusion strategies (e.g., bilinear pooling \cite{lin2015bilinear, kim2016hadamard}, cross-attention \cite{wei2020multi}, gating\cite{ren2018gated}) to reduce confounding factors and keep the benchmark focused on representation quality. Richer fusion operators are an orthogonal direction for future work.

\paragraph{Classification Head.}
The fused feature vector \(z\), obtained by combining representations from all available branches (chart-based and optionally numerical) via the fusion module, is passed to a fully connected classification head that learns non-linear decision boundaries. This head consists of one or more dense layers with ReLU activations and dropout regularization to mitigate overfitting, followed by a final softmax layer that outputs class probabilities:
\[
\hat{y} = \mathrm{softmax}\big(W_2\,\sigma(W_1 z + b_1) + b_2\big)
\]
where \(W_1, W_2\) and \(b_1, b_2\) are learnable and trainable parameters, and \(\sigma(\cdot)\) denotes a standard non-linear activation (e.g., ReLU). The dimensionality of \(z\) depends on the selected fusion strategy and number of modalities, allowing the classifier to adapt seamlessly across single-chart, multi-chart, and multimodal configurations.

\paragraph{Multichart-Multimodal Architecture.}
Let \(\mathcal{I}_k(x_i)\) represent the \(k\)-th chart image (e.g., line, bar) derived from a raw time series \(x_i\), and let \(x_i^{\text{num}}\) denote the corresponding numerical sequence. The numerical branch is processed through its dedicated encoder:
\[
h_{\text{num}} = \phi_{\text{num}}(x_i^{\text{num}})
\]
while each chart passes through its modality-specific CNN encoder:
\[
h_k = \phi_k(\mathcal{I}_k(x_i)) \quad \text{for } k = 1, \dots, K
\]
The resulting representations are fused using a mechanism:
\begin{itemize}[leftmargin=10pt]
    \item \textbf{Multi-chart only:} \( z = \text{Fusion}(h_1, \dots, h_K) \)
    \item \textbf{Multimodal:} \( z = \text{Fusion}(h_{\text{num}},\, h_1, \dots, h_K) \)
\end{itemize}
The fused vector \(z\) is then passed to the classification head for label prediction. Unlike prior image-based TSC approaches that rely on a single chart projection, VTBench leverages multiple chart perspectives alongside the raw signal, allowing the model to access complementary cues—visual shape cues across projections and precise temporal measurements from the numerical branch. 
To mitigate overfitting and negative transfer from noisy modalities, we use (i) lightweight CNN backbones to limit branch capacity relative to data size, and (ii) a learnable attention-based fusion mechanism to down-weight uninformative views, preventing any single chart from dominating predictions. The use of independent, parallelizable encoders ensures scalability, enabling easy addition or removal of modalities without altering the overall architecture. This design offers a principled way to leverage complementary, noise-robust, and dataset-adaptive multiview information.


\paragraph{Supported Configurations.}
\syss is designed to support a range of model configurations via a unified and extensible interface:
\begin{itemize}[leftmargin=10pt]
    \item \textbf{Single-Chart:} A unimodal CNN applied to a single chart type.
    \item \textbf{Multi-Chart:} Parallel CNN branches for all chart types with feature-level fusion across visual modalities.
    \item \textbf{Multimodal:} A combination of multi-chart CNNs and a numerical branch to jointly model visual and temporal information.
\end{itemize}
This modular design allows plug-and-play experimentation across architecture types, enabling reproducible ablation studies and comprehensive benchmark comparisons under consistent settings.

\section{Experiments}
\subsection{Experimental Setup}

To rigorously evaluate the effectiveness of chart-based visual encodings and multimodal fusion for time-series classification (TSC), we conduct systematic experiments using \syss across diverse UCR datasets, charting choices, and fusion strategies. Our goal is to quantify the contribution of each component—chart type, visual design choices, and numerical signal—in isolation and in combination.

\paragraph{Datasets.}
We evaluate our framework on 31 benchmark datasets from the UCR Time Series Archive \cite{dau2019ucr}, which span a wide range of sequence lengths, class cardinalities, and application domains (see Table~\ref{tab:dataset_char_ablation} for full details). This diversity allows us to assess whether chart-based representations generalize across varied temporal patterns. Both binary and multiclass classification tasks are included to capture differences in modality effects across task complexity. We adopt the standard train/test splits provided by the archive and extract a stratified 80/20 split from the test set to form a validation set, ensuring class balance throughout.

\paragraph{Chart Generation and Ablation Factors.}
Each univariate time-series instance is converted into four standard visual encodings—\textit{line}, \textit{bar}, \textit{area}, and \textit{scatter} plots—using \texttt{matplotlib}. To assess how visual characteristics affect classification, we systematically vary three key parameters: (1) \textbf{Chart type}, reflecting different temporal structures; (2) \textbf{Visual styling}, including color mode (RGB vs.\ grayscale) and axis annotations (presence or absence of tick marks, titles, and bounding boxes); and (3) \textbf{Image resolution}, with candidate sizes of 64$\times$64, 128$\times$128, and 256$\times$256. Unless otherwise stated, charts are rendered at 128$\times$128 for consistency across datasets.

These visual encoding choices, along with model-level configurations, constitute the core of our ablation study. Specifically, we vary the following dimensions:
\begin{enumerate}[leftmargin=1.5em, topsep=2pt]
    \item \textbf{Chart type:} Line, bar, area, and scatter plots.
    \item \textbf{Visual settings:} Color mode and axis label visibility.
    \item \textbf{Image resolution:} 64$\times$64, 128$\times$128, 256$\times$256 (see Appendix ~\ref{sec:appendix:ablation_study}).
    \item \textbf{Fusion strategy:} Feature concatenation vs.\ attention-based weighted fusion.
    \item \textbf{Numerical encoder:} Omni-Scale CNN vs.\ Transformer-based architectures.
\end{enumerate}
This design enables controlled comparison of component-level contributions to overall performance. A detailed analysis of the empirical effects of these choices is presented in Subsection~\ref{subsec:results}.

\paragraph{Training Details.}
All CNN branches are trained from scratch using the Adam optimizer with a learning rate of $10^{-3}$ and weight decay of $10^{-2}$. We apply early stopping with a patience of 10 epochs based on validation accuracy to prevent overfitting. A \textit{ReduceLROnPlateau} scheduler halves the learning rate if validation loss fails to improve over 3 consecutive epochs. The cross-entropy loss function is used for all configurations. Training is conducted on RTX 4090 or RTX 6000 Ada GPUs, with every single multimodal run per dataset completing in under 5 minutes. For unimodal (single-chart) models, we report \textbf{mean accuracy} across 10 independent runs using different random seeds to account for optimization variability. For multimodal models, we report the average over three independent runs, as pilot experiments showed negligible variance (<0.3\%) across runs, making repeated sweeps \mbox{computationally unnecessary}.

\setlength{\tabcolsep}{1.5pt}
\renewcommand{\arraystretch}{0.80}
\begin{table*}[!t]
\centering
\scriptsize
\caption{\textbf{Performance comparison of VTBench against SOTA time-series classifiers across selected UCR datasets.} 
\textbf{Best VTBench} refers to the highest-performing configuration within VTBench that uses chart-based representations, either as a single-chart model, a 2-branch fusion (chart + numerical input), or a 5-branch fusion. 
\textbf{Default VTBench} denotes the full multimodal configuration that fuses all chart types with numerical input with consistent chart and model settings. Best results per dataset are highlighted in bold, and second-best results are underlined.}
\label{tab:vtbench_vs_sota}
\begin{tabular}{c|c|c|c|c|c|c|c|c|c|c|c|c|c}
\toprule
\multirow{2}{*}{\textbf{Task}} &
\multirow{2}{*}{\textbf{Dataset}} &
\multirow{2}{*}{\textbf{COTE}} &
\multirow{2}{*}{\textbf{HC2}} &
\multirow{2}{*}{\textbf{OS-CNN}} &
\multirow{2}{*}{\textbf{FCN-LSTM}} &
\multirow{2}{*}{\textbf{IncTime}} &
\multirow{2}{*}{\textbf{ROCKET}} &
\multirow{2}{*}{\textbf{CHIEF}} &
\multirow{2}{*}{\textbf{TimesNet}} &
\multirow{2}{*}{\textbf{GPT4TS}} &
\multirow{2}{*}{\textbf{SoftShape}} &
\textbf{Default} &
\textbf{Best} \\
& & & & & & & & & & & & \textbf{VTBench} & \textbf{VTBench} \\
\midrule

\multirow{18}{*}{\textbf{Binary}} 
& StrawBerry & 0.963 & 0.979 & 0.982 & 0.984 & 0.984 & 0.981 & 0.968 & 0.977 & 0.974 & \textbf{0.989} & 0.954 & 0.964 \\
& Yoga & 0.898 & 0.930 & 0.911 & 0.918 & 0.906 & 0.910 & 0.848 & \underline{0.954} & 0.948 & \textbf{0.983} & 0.813 & 0.829 \\
& FordB & 0.963 & 0.930 & 0.838 & 0.918 & 0.937 & 0.805 & 0.832 & \underline{0.905} & 0.929 & \textbf{0.974} & 0.777 & 0.788 \\
& GunPoint & 0.991 & \underline{0.999} & \underline{0.999} & \textbf{1.000} & \textbf{1.000} & \textbf{1.000} & \textbf{1.000} & 0.985 & 0.980 & 0.990 & 0.911 & 0.971 \\
& Lightning2 & 0.785 & 0.783 & 0.861 & 0.803 & 0.803 & 0.759 & 0.771 & \underline{0.894} & \textbf{0.895} & 0.886 & 0.694 & 0.781 \\
& ItalyPowerDemand & 0.970 & 0.962 & 0.947 & 0.963 & 0.965 & 0.970 & 0.972 & \textbf{0.986} & \underline{0.982} & \underline{0.982} & 0.909 & 0.960 \\
& PhalangesOutlineCorrect & 0.783 & 0.845 & 0.830 & 0.837 & 0.850 & 0.845 & 0.825 & \textbf{0.932} & 0.855 & \underline{0.930} & 0.781 & 0.801 \\
& Wafer & \underline{0.999} & \textbf{1.000} & 0.998 & \underline{0.999} & \underline{0.999} & 0.998 & \underline{0.999} & 0.998 & \underline{0.999} & \textbf{1.000} & 0.996 & 0.996 \\
& Wine & 0.904 & \textbf{0.925} & 0.744 & 0.870 & 0.887 & \underline{0.922} & 0.898 & 0.829 & 0.856 & 0.884 & 0.512 & 0.549 \\
& ToeSegmentation1 & 0.934 & 0.955 & 0.954 & \underline{0.974} & 0.953 & 0.932 & 0.960 & 0.851 & 0.885 & \textbf{0.985} & 0.769 & 0.842 \\
& ToeSegmentation2 & 0.952 & 0.960 & 0.946 & 0.962 & \underline{0.964} & 0.933 & 0.963 & 0.832 & 0.814 & \textbf{0.988} & 0.800 & 0.771 \\
& SonyAIBORobotSurface1 & 0.899 & 0.952 & 0.980 & 0.985 & 0.954 & 0.958 & 0.890 & \textbf{0.995} & 0.989 & \underline{0.994} & 0.601 & 0.683 \\
& Ham & 0.805 & 0.859 & 0.704 & 0.781 & 0.714 & 0.857 & 0.805 & 0.902 & \textbf{0.935} & \underline{0.916} & 0.686 & 0.742 \\
& Beetlefly & 0.921 & 0.903 & 0.815 & 0.950 & 0.893 & 0.887 & \underline{0.958} & 0.700 & 0.725 & \textbf{0.975} & 0.700 & 0.833 \\
& Computers & 0.770 & 0.859 & 0.707 & 0.860 & \underline{0.866} & 0.842 & 0.754 & 0.812 & 0.808 & 0.816 & 0.836 & \textbf{0.875} \\
& Herring & 0.632 & 0.626 & 0.608 & 0.766 & 0.625 & 0.626 & 0.597 & 0.656 & \underline{0.800} & 0.767 & 0.573 & 0.599 \\
& Earthquakes & 0.747 & 0.748 & 0.670 & 0.835 & 0.731 & 0.749 & 0.748 & \textbf{0.914} & \underline{0.905} & 0.892 & 0.739 & 0.765 \\

\midrule

\multirow{14}{*}{\textbf{Multiclass}} 
& Adiac & 0.810 & 0.795 & 0.835 & 0.859 & \underline{0.863} & 0.806 & 0.780 & 0.808 & 0.807 & \textbf{0.937} & 0.652 & 0.663 \\
& ArrowHead & 0.874 & 0.886 & 0.838 & \underline{0.909} & 0.880 & 0.854 & 0.881 & 0.782 & 0.835 & \textbf{0.944} & 0.602 & 0.739 \\
& Beef & 0.764 & 0.797 & 0.807 & 0.900 & 0.682 & 0.751 & 0.632 & 0.800 & \textbf{0.933} & 0.867 & 0.867 & \underline{0.911} \\
& ChlorineConcentration & 0.736 & 0.769 & 0.839 & 0.810 & 0.864 & 0.794 & 0.661 & \textbf{0.999} & \textbf{0.999} & \textbf{0.999} & 0.624 & 0.676 \\
& Crop & - & 0.764 & 0.769 & - & 0.772 & 0.752 & 0.762 & 0.850 & \underline{0.876} & \textbf{0.877} & 0.737 & 0.726 \\
& CricketX & 0.814 & 0.840 & 0.821 & 0.808 & \underline{0.853} & 0.841 & 0.830 & 0.797 & 0.791 & \textbf{0.932} & 0.696 & 0.677 \\
& CricketY & 0.815 & 0.850 & \underline{0.867} & 0.818 & 0.860 & 0.845 & 0.817 & 0.781 & 0.803 & \textbf{0.915} & 0.669 & 0.694 \\
& CricketZ & 0.827 & 0.859 & \underline{0.863} & 0.810 & 0.861 & 0.852 & 0.838 & 0.796 & 0.803 & \textbf{0.915} & 0.724 & 0.737 \\
& ECG5000 & 0.963 & 0.948 & 0.940 & 0.947 & 0.941 & 0.947 & 0.946 & 0.971 & \underline{0.975} & \textbf{0.979} & 0.930 & 0.938 \\
& FaceAll & \underline{0.990} & 0.987 & 0.845 & 0.940 & 0.983 & 0.989 & 0.983 & 0.973 & 0.988 & \textbf{0.998} & 0.738 & 0.841 \\
& FacesUCR & 0.967 & 0.975 & 0.967 & 0.929 & 0.977 & 0.972 & 0.973 & \underline{0.980} & 0.979 & \textbf{0.995} & 0.859 & 0.869 \\
& InsectWingbeat & 0.639 & 0.661 & 0.635 & 0.653 & 0.627 & 0.656 & 0.632 & 0.736 & \underline{0.852} & \textbf{0.888} & 0.702 & 0.704 \\
& RefrigerationDevices & 0.742 & \underline{0.767} & 0.503 & 0.581 & 0.759 & 0.726 & 0.727 & 0.569 & 0.727 & \textbf{0.803} & 0.631 & 0.612 \\
& WordSynonyms & 0.748 & 0.766 & 0.742 & 0.671 & 0.752 & 0.766 & 0.794 & \underline{0.844} & 0.782 & \textbf{0.918} & 0.645 & 0.658 \\

\midrule
\multicolumn{2}{c|}{\textbf{Avg. Accuracy}} & 0.851 & 0.867 & 0.831 & 0.868 & 0.861 & 0.856 & 0.840 & 0.865 & 0.885 & 0.933 & 0.746 & 0.780 \\
\bottomrule
\end{tabular}
\end{table*}
\subsection{Results Analysis}
\label{subsec:results}

We evaluated \syss against a broad spectrum of state-of-the-art (SOTA) time-series classification (TSC) models spanning diverse methodological paradigms (Table~\ref{tab:vtbench_vs_sota}): an ensemble-based method HC2~\cite{middlehurst2021hive}, deep CNN and hybrid architectures IncTime~\cite{ismail2020inceptiontime}, a random convolution kernel approach ROCKET~\cite{dempster2020rocket}, a tree-based ensemble CHIEF~\cite{shifaz2020ts}, a general-purpose time-series transformer TimesNet~\cite{TimesNet}, a foundation model GPT4TS~\cite{zhou2023one}, and a recent shapelet-based model SoftShape~\cite{liu2025learning}. 
This comprehensive comparison situates \syss within the wider TSC landscape.

Overall, the \textbf{Best \sys} configuration (single or multi-chart with or without numerical input) achieved an average accuracy of 78.0\%, remaining below leading methods such as TimesNet (86.5\%), HC2 (86.7\%), and GPT4TS (88.5\%), as well as the specialized SoftShape model (93.4\%). Rather than competing directly with these methods, \syss highlights settings where chart-based and multimodal representations provide consistent and interpretable gains across datasets. The \textbf{default \sys} (full fusion variant)—using default parameters informed by generalizable empirical insights—achieved an average accuracy of 73.0\%, also below top-performing baselines overall, but with notable variation across datasets.

Performance differed significantly by dataset. For example, on GunPoint, the best single-modality configuration reached 0.971 accuracy, compared to 0.911 for full fusion. In contrast, on Herring, the full multimodal setup yielded better results. These outcomes suggest that chart-based features can meaningfully complement raw numerical inputs, but including all possible chart modalities may introduce redundancy or noise. Below, we revisit the five key research questions introduced earlier in Section~\ref{intro} to clarify the conditions under which chart-based representations are most effective for TSC.



\subsubsection{Chart Type Effectiveness Under Different Conditions}

\begin{table}[!ht]
\centering
\small
\renewcommand{\arraystretch}{0.8}
\caption{\textbf{Aggregate accuracy (\%) for each chart type grouped by task type under four rendering settings.}
Values are mean ± 95\% CI across datasets. \textbf{Bold} = best per row. * indicates highest accuracy per group.}
\label{tab:agg_chart_task_type}
\begin{tabular}{llcccc}
\toprule
\textbf{Group} & \textbf{Chart} &
\multicolumn{2}{c}{\textbf{Monochrome}} &
\multicolumn{2}{c}{\textbf{Color}} \\
\cmidrule(lr){3-4}\cmidrule(lr){5-6}
& &
\textbf{Label} & \textbf{No Label} &
\textbf{Label} & \textbf{No Label} \\
\midrule
\multirow{4}{*}{Binary} 
& Line    & \textbf{78.4 ± 6.6*} & 77.3 ± 7.5 & 75.5 ± 8.9 & 75.5 ± 8.5 \\
& Area    & 74.2 ± 8.2 & \textbf{76.9 ± 7.0} & 74.5 ± 7.6 & 75.3 ± 8.1 \\
& Bar     & 74.7 ± 7.4 & 75.3 ± 8.2 & 74.5 ± 7.9 & \textbf{76.2 ± 7.1} \\
& Scatter & 75.0 ± 7.6 & 76.0 ± 7.5 & 75.7 ± 7.2 & \textbf{76.3 ± 7.3} \\
\midrule
\multirow{4}{*}{Multiclass} 
& Line    & 64.6 ± 5.8 & \textbf{67.2 ± 5.7} & 64.2 ± 6.1 & 64.5 ± 8.0 \\
& Area    & \textbf{66.5 ± 5.4} & 65.5 ± 6.3 & 64.8 ± 5.7 & 65.3 ± 6.5 \\
& Bar     & 64.4 ± 5.9 & 65.6 ± 6.5 & 64.0 ± 6.0 & \textbf{65.9 ± 5.3} \\
& Scatter & 66.3 ± 6.1 & \textbf{67.5 ± 5.9*} & 65.1 ± 6.0 & 66.1 ± 6.3 \\
\bottomrule
\end{tabular}
\end{table}

Our input-level ablation (Table~\ref{tab:agg_chart_task_type}) compared four common chart types---\textbf{line}, \textbf{area}, \textbf{bar}, and \textbf{scatter}---rendered as 128$\times$128 images.

\parasum{No single chart format universally dominates; the optimal choice is context-dependent.}
Continuous trajectory charts, particularly line plots, consistently yielded the highest classification accuracy across time-series tasks. In \syss multimodal evaluations, line charts achieved up to 78.4\% accuracy, outperforming bar (76.2\%) and scatter (76.3\%) plots on average. This benefit arises from their ability to encode smooth temporal transitions, allowing convolutional filters to capture local structure and shape variations. Area charts, as filled variants of line plots, offered similar advantages but showed higher variance across datasets. In contrast, bar and scatter plots discretize the signal into bins or points, disrupting temporal continuity and reducing feature extraction quality. These results suggest that preserving the continuous nature of time-series data in visual encodings is key for maximizing discriminative utility.

\parasum{The performance ranking of chart types shifts with data characteristics.}
The advantage of continuous trajectory charts diminishes with long sequences, particularly when the $Timesteps \over Pixels$ ratio is high (Table~\ref{tab:dataset_char_ablation}). In such cases, overplotting leads to visual saturation, obscuring fine-grained patterns and compressing critical temporal variation. All chart types then exhibit similar degradation in accuracy. Moreover, visual design choices such as removing axis labels can reduce clutter and improve performance in simpler tasks, while color encoding can help distinguish class-relevant features. Although these stylistic factors are secondary to chart type, they can further modulate effectiveness depending on the dataset.

\parasum{Practical takeaways.}
Line or area charts are generally effective for short-to-moderate-length sequences, where temporal transitions remain visually distinct. For longer sequences, compression effects can reduce the utility of continuous charts, and simpler chart types may perform comparably. Reducing clutter and incorporating class-relevant visual cues, such as color, can help maintain performance under these conditions.

\setlength{\tabcolsep}{1.5pt}
\renewcommand{\arraystretch}{0.75}
\begin{table*}[!t]
\centering
\scriptsize
\caption{\textbf{Comparison of VTBench with the best-performing single-chart configuration across datasets.} 
Results are grouped into \textit{Improving}, \textit{Almost Same}, and \textit{Degrading} based on $\Delta = \text{VTBench} - \text{Single}$.}

\label{tab:dataset_char_ablation}
\begin{tabular}{c|c|c|c|c|c|c|c|c|c|c|c}
\toprule
\multirow{2}{*}{\textbf{Task Type}} &
\multirow{2}{*}{\textbf{Dataset}} &
\multirow{2}{*}{\textbf{\#Timesteps}} &
\multirow{2}{*}{\textbf{\#Train}} &
\multirow{2}{*}{\textbf{\#Test}} &
\multirow{2}{*}{\textbf{Domain}} &
\multirow{2}{*}{\textbf{Type}} &
\multirow{2}{*}{\textbf{Best Chart Type}} &
\multirow{2}{*}{\textbf{Best Setting}} &
\textbf{Best Single} &
\textbf{Default} &
\textbf{$\Delta$ VTBench} \\
& & & & & & & & & \textbf{Chart} & \textbf{VTBench} &
\textbf{(VTBench $-$ Single)} \\
\midrule

\multicolumn{12}{c}{\textbf{Improving Performance}} \\
\midrule
Multiclass & Beef & 470 & 30 & 30 & Food Sci & Spectro & Bar & Color-L & 0.588 & 0.867 & +0.288 \\
Multiclass & InsectWingbeatSound & 256 & 220 & 1980 & Biology & Audio & Area & Color-L & 0.653 & 0.702 & +0.049 \\
Binary & Computers & 720 & 250 & 250 & Energy & Device & Line & Color-L & 0.704 & 0.836 & +0.132 \\
Multiclass & RefrigerationDevices & 720 & 375 & 375 & Energy & Device & Line & Color-NL & 0.543 & 0.631 & +0.088 \\
Multiclass & Adiac & 176 & 390 & 391 & Biology & Image & Line & Color-NL & 0.631 & 0.652 & +0.021 \\

\midrule

\multicolumn{12}{c}{\textbf{Almost Same Performance}} \\
\midrule
Binary & FordB & 500 & 3636 & 810 & Automotive & Sensor & Area & Mono-NL & 0.782 & 0.777 & -0.005 \\
Binary & Wafer & 152 & 1000 & 6164 & Manufacturing & Sensor & Bar & Color-L & 0.997 & 0.996 & -0.001 \\
Binary & Ham & 431 & 100 & 105 & Food Sci & Spectro & Area & Color-NL & 0.703 & 0.686 & -0.017 \\
Multiclass & ECG5000 & 140 & 500 & 4500 & Healthcare & Sensor & Scatter & Mono-NL & 0.937 & 0.930 & -0.007 \\
Multiclass & CricketZ & 300 & 390 & 390 & Motion & Human Activity & Line & Mono-NL & 0.755 & 0.724 & -0.031 \\
Binary & Earthquakes & 512 & 322 & 139 & Geophysics & Sensor & Line & Mono-L & 0.738 & 0.739 & +0.001 \\
Binary & StrawBerry & 235 & 613 & 370 & Food Sci & Spectro & Line & Mono-NL & 0.976 & 0.954 & -0.022 \\
Binary & Herring & 512 & 64 & 64 & Biology & Image & Scatter & Mono-NL & 0.598 & 0.573 & -0.025 \\
\midrule

\multicolumn{12}{c}{\textbf{Degrading Performance}} \\
\midrule
Binary & Wine & 234 & 57 & 54 & Chemistry & Spectro & Line & Color-NL & 0.751 & 0.512 & -0.239 \\
Multiclass & ChlorineConcentration & 166 & 467 & 3840 & Chemistry & Simulation & Line & Mono-L & 0.752 & 0.624 & -0.128 \\
Multiclass & FaceAll & 131 & 560 & 1690 & Biometrics & Image & Area & Mono-NL & 0.821 & 0.738 & -0.083 \\

Binary & ItalyPowerDemand & 24 & 67 & 1029 & Energy & Sensor & Line & Color-NL & 0.969 & 0.909 & -0.060 \\
Binary & PhalangesOutlinesCorrect & 80 & 1800 & 858 & Healthcare & Image & Line & Mono-NL & 0.817 & 0.781 & -0.036 \\
Multiclass & Crop & 46 & 7200 & 16800 & Agriculture & Image & Area & Mono-NL & 0.776 & 0.737 & -0.039 \\
Multiclass & WordSyn & 270 & 267 & 638 & NLP & Image & Scatter & Mono-L & 0.681 & 0.645 & -0.036 \\
Binary & Beetlefly & 512 & 20 & 20 & Biology & Image & Line & Mono-L & 0.925 & 0.700 & -0.225 \\
Multiclass & CricketY & 300 & 390 & 390 & Motion & Human Activity & Line & Color-L & 0.730 & 0.669 & -0.061 \\
Multiclass & Arrowhead & 251 & 36 & 175 & Anthropology & Image & Line & Color-L & 0.791 & 0.602 & -0.189 \\
Binary & Yoga & 426 & 300 & 3000 & Motion & Image & Area & Mono-NL & 0.876 & 0.813 & -0.063 \\
Binary & GunPoint & 150 & 50 & 150 & Motion & Human Activity & Line & Mono-NL & 0.997 & 0.911 & -0.086 \\
Binary & Lightning2 & 637 & 60 & 61 & Weather & Sensor & Area & Color-NL & 0.846 & 0.694 & -0.152 \\
Binary & ToeSegmentation1 & 277 & 40 & 228 & Motion & Motion & Line & Color-L & 0.921 & 0.769 & -0.152 \\
Binary & ToeSegmentation2 & 343 & 36 & 130 & Motion & Motion & Line & Mono-NL & 0.938 & 0.800 & -0.138 \\
Binary & SonyAIBORobotSurface1 & 70 & 20 & 601 & Robotics & Sensor & Line & Mono-L & 0.831 & 0.601 & -0.230 \\
Multiclass & CricketX & 300 & 390 & 390 & Motion & Human Activity & Line & Color-NL & 0.750 & 0.696 & -0.054 \\

Multiclass & FacesUCR & 131 & 200 & 2050 & Biometrics & Image & Area & Color-L & 0.952 & 0.859 & -0.093 \\

\bottomrule
\end{tabular}
\end{table*}

\subsubsection{Utility Across Domains and Data Characteristics}
Table~\ref{tab:dataset_char_ablation} shows that the effectiveness of chart-based representations is highly domain- and data-dependent, with a noticeable contrast between non-motion and motion-oriented time-series tasks.

\parasum{Non-motion domains often benefit from chart visualizations, particularly when fused with raw data}, such as biological signals, chemical spectra, or food science.
These datasets frequently exhibit discriminative local shapes, peaks, or oscillatory patterns that are naturally emphasized in static images. Multimodal models improved classification accuracy in several such cases (e.g., device datasets like Computers and RefrigerationDevices), suggesting that charts can act as complementary feature extractors that make salient waveform structures more accessible to convolutional encoders.

\parasum{Motion-oriented datasets more frequently showed performance degradation when chart modalities were added}, such as human activity recognition, gesture, or robotics sensor streams. These tasks rely heavily on fine-grained temporal ordering and phase information, which static visual snapshots cannot fully preserve. \syss experiments indicate that introducing multiple chart views in these domains can inject redundant or misaligned cues, effectively adding noise and lowering accuracy (e.g., GunPoint and ToeSegmentation datasets), although a few cases remain stable.

\parasum{Impacts of number of classes matter.} Multimodal configurations tend to be more beneficial (or at least not harmful) for binary classification problems than for multiclass ones. Binary tasks often showed improvements or stability when visual modalities were added, likely because separating two classes can benefit from additional discriminative cues. In multiclass settings, however, the added charts sometimes made it harder for the model to distinguish among many classes, especially when redundant patterns or spurious variations were introduced.

\parasum{Practical takeaways.}
Chart-based representations are most effective for domains where local shapes or intensity profiles define class boundaries, and they are particularly useful when fused with raw data. For highly sequential domains, where phase and temporal ordering are critical, static visualizations may degrade performance and should be applied with caution. Performance should be monitored closely as task complexity increases, and visual modalities may require selective inclusion or tuning to avoid redundancy or misalignment.


\subsubsection{Charts as Complementary (Not Replacement) to Raw Inputs}
A central question is whether charts should augment the original time-series data or replace it as the sole input modality. \sys's empirical results (Appendix Table \ref{tab:charts_numerical_ablation}) indicate that chart-based visualizations are best used as a complement to raw inputs rather than as a standalone replacement.

\parasum{Including the numerical series alongside chart features consistently improved or maintained accuracy over chart-only models.} Best-performing \syss configurations fused raw and visual features, with charts highlighting shape- and pattern-level cues while raw inputs preserved precise timing and values. Across many datasets, multimodal fusion (raw + charts) often achieved top accuracy, suggesting that charts revealed visual motifs not fully captured by numerical data alone.

\parasum{Complementarity of charts is not unconditional.}
In some settings, especially motion-based or multiclass tasks, charts introduced redundant or noisy features that degraded performance. This occurred when the charts failed to add new information beyond what the raw series already provided. Binary tasks were generally more robust, with charts often helping or having minimal downside. Complementary value arises only when the added visual signals provide non-overlapping cues. 

\parasum{Among fusion strategies, \textbf{dynamic weighted fusion} with a Transformer-based encoder was most effective,}
achieving 82.0\% mean accuracy and ranking first on 8 of 12 benchmark datasets ( Table~\ref{tab:model_ablation}). Adaptive fusion methods outperformed fixed strategies like simple concatenation, highlighting the value of letting the model learn which modality to prioritize.

\parasum{Practical takeaways.}
Use charts to augment, not replace, raw time-series inputs when the visual patterns provide complementary information. Multimodal setups combining multiple chart views with raw data tend to perform well, especially with adaptive fusion mechanisms like dynamic weighting. If performance degrades, it may indicate that the charts are redundant or noisy; in such cases, fusion strategies that can down-weight unhelpful inputs are preferred. Effective multimodal models should allow selective use of charts when they offer added value, while relying more on raw data when they do not. 

\begin{table}[!t]
\centering
\small
\caption{\textbf{Aggregate mean accuracy (\%) across 12 datasets} comparing fusion strategies and numerical encoders for VTBench multimodal configurations (128×128 resolution, DeepCNN backbone).}
\label{tab:model_ablation}
\begin{tabular}{lcccc}
\toprule
\makecell{\textbf{Numerical}\\\textbf{Encoder}} & 
\makecell{\textbf{Fusion}\\\textbf{Strategy}} & 
\makecell{\textbf{Mean}\\\textbf{Accuracy}} & 
\makecell{\textbf{Std.}\\\textbf{Dev.}} & 
\makecell{\textbf{Best}\\\textbf{Rank Count}} \\
\midrule
Transformer & Concatenation         & 0.807 & \textbf{11.01} & 3/12 \\
\textbf{Transformer} & \textbf{Weighted Fusion} & \textbf{0.820} & 11.87 & \textbf{8/12} \\
OS-CNN      & Concatenation         & 0.810 & 12.84 & 2/12 \\
OS-CNN      & Weighted Fusion       & 0.806 & 12.17 & 3/12 \\
\bottomrule
\end{tabular}
\vspace{-2em}
\end{table}

\subsubsection{Impact of Training Set Size and Sequence Length}
\syss performance is shaped by both sequence length and dataset size. 

\parasum{On smaller datasets, chart-only models can be selectively effective, but multimodal fusion is fragile.}  
Multimodal models risk overfitting on limited data due to their larger parameter count. Simple charts—monochrome, no axis labels—performed best by minimizing distractions. Visual enhancements like color gradients helped, but extraneous details (e.g., tick marks, text) hurt generalization.

\parasum{Larger datasets benefit from added modalities.}  
With sufficient training samples, models learned to exploit complementary features from both modalities. In these cases, full multimodal setups outperformed single-input baselines, as data volume reduced sensitivity to noise and redundancy (Table \ref{tab:dataset_char_ablation}).

\parasum{Sequence length interacts with chart resolution.}  
Shorter sequences (under 200 timesteps) were well captured at 128$\times$128 resolution (Table \ref{tab:image_size_stats}). Lower resolutions (e.g., 64$\times$64) lost detail, while higher ones (256$\times$256) brought no consistent gains. For long sequences, high $Timesteps \over Pixels$ ratios caused overplotting and visual compression. Increasing resolution had diminishing returns unless the model architecture scaled accordingly. Segmenting long series into subcharts or combining chart views with sequence models could help, but could also introduce redundancy.

\parasum{Practical takeaways.}  
Use minimalistic yet informative charts for small datasets; avoid clutter and emphasize core patterns (e.g., via color gradients). Multimodal fusion is more reliable with larger datasets where overfitting is less likely. For short-to-moderate sequences, 128$\times$128 charts offer a good trade-off between fidelity and efficiency. For very long sequences, consider segmenting input or incorporating sequence-aware models. Avoid excessive chart resolution unless model capacity is scaled to match, and visually inspect chart clarity to ensure that key features are not obscured.

\subsubsection{Unique Advantages of Visual Chart Representations over Traditional Approaches}
Chart-based time-series representations offer several distinct benefits over purely numerical approaches:

\parasum{Human-aligned, interpretable features.}  
Charts mirror how humans analyze time-series---through visual patterns. This alignment enhances interpretability, especially in domains like healthcare or finance. Chart-based models capture visual motifs (e.g., slopes, spikes, oscillations) that convolutional encoders can detect directly, often more easily than raw-value models.

\parasum{Reuse of vision models.}  
Converting time-series into images enables direct use of pre-trained CNNs and Vision Transformers. This supports faster development, transfer learning, and modular model design. Unlike specialized time-series models, visual pipelines can leverage the broader vision ecosystem with minimal adaptation.

\parasum{Multi-view feature complementarity.}  
Multiple chart types (line, area, bar, scatter) can be easily generated to highlight different temporal properties. Their fusion yields richer representations without manual feature engineering. \syss showed consistent gains from this multi-chart setup, demonstrating its flexibility \mbox{and effectiveness}.

\parasum{Robustness and generalization.}  
Chart+raw models showed stable performance across diverse datasets, avoiding extreme failures seen in some baselines. This robustness stems from multimodal redundancy—the model can rely on either visual or numerical inputs depending on the case—making it more dependable in real-world, variable scenarios.

\parasum{Stronger performance in low-data settings.}  
On small datasets, charts act as implicit feature extractors, helping models generalize with limited data. CNNs on charts often matched or exceeded raw-based models, aided by visual smoothing and built-in inductive bias. Visual augmentation (e.g., color shifts, scale jittering) also enhances generalization with minimal overhead.

\parasum{Practical takeaways.}  
Chart representations offer interpretability, leverage existing vision models, support multi-view fusion, and improve robustness—especially in low-data regimes. They provide a general-purpose, architecture-agnostic foundation for time-series modeling with strong out-of-the-box performance.

\begin{table}[!t]
\centering
\small
\caption{\textbf{Average classification accuracy and Wilcoxon signed-rank test results across image sizes.} 
Accuracy values are averaged over datasets grouped by time-series length. The last row shows the overall mean accuracy $\pm$ half-width of the 95\% confidence interval. Statistical test results report p-values and effect sizes ($\alpha = 0.05$), where a \ding{51} indicates significance and Cliff's $\delta$ quantifies effect magnitude (small $\approx$ 0.1, medium $\approx$ 0.33, large $\approx$ 0.47).}
\vspace{-1em}
\label{tab:image_size_stats}
\begin{tabular}{lccc}
\toprule
\textbf{Length Group} & \textbf{64×64} & \textbf{128×128} & \textbf{256×256} \\
\midrule
Short (<200)     & 0.8105 & 0.8295 & 0.8486 \\
Medium (200–400) & 0.7693 & 0.7532 & 0.7381 \\
Long (>400)      & 0.7135 & 0.7298 & 0.7263 \\
\midrule
\textbf{Overall}\textsuperscript{†} 
& 0.7641 $\pm$ 0.046 
& 0.7720 $\pm$ 0.041 
& 0.7731 $\pm$ 0.044 \\
\midrule
\multicolumn{4}{l}{\textbf{Wilcoxon signed-rank test results with effect sizes}} \\
\midrule
\textbf{Comparison} & \textbf{p-value} & \textbf{Significant?} & \textbf{Cliff's $\delta$} \\
128$\times$128 vs 64$\times$64   & 0.0086   & \textbf{\ding{51}} & 0.0604 \\
256$\times$256 vs 128$\times$128 & 0.6919   & \textbf{\ding{55}} & -0.0052 \\
256$\times$256 vs 64$\times$64   & 0.0606   & \textbf{\ding{55}} & 0.0510 \\
\bottomrule
\end{tabular}
\vspace{-1em}
\end{table}

\section{Conclusion}
This paper introduces \sys, a framework for time-series classification that reimagines temporal signals as interpretable visual representations. By integrating chart-based views with raw numerical inputs, \syss supports both multiview and multimodal configurations that capture diverse and complementary temporal features. Empirical results show that these configurations can improve or maintain performance over single-chart baselines on selected UCR datasets, while also revealing cases where additional modalities introduce redundancy. \syss further examines the interplay between image resolution and sequence length, providing practical guidance for chart design. While it does not surpass specialized SOTA models, \syss achieves competitive performance in selective datasets, strong generalization behavior, and a modular architecture that facilitates extension to additional modalities such as frequency-domain features.

In summary, visual chart representations enhance time-series classification by improving interpretability, enabling reuse of vision models, and supporting flexible multimodal learning. \syss shows that when aligned with data characteristics, chart-based inputs can provide complementary information, though their effectiveness is dataset-dependent. Beyond accuracy, \textbf{\syss offers a systematic empirical foundation}---highlighting how factors such as chart type, sequence length, image resolution, and training set size interact to shape model behavior. These insights establish practical design principles and open avenues for future research, including adaptive chart generation, integration with domain-specific visual encodings, and automated modality selection. As such, \syss lays the groundwork for more interpretable, flexible, and data-aware time-series modeling frameworks.

\clearpage
\bibliographystyle{ACM-Reference-Format}
\bibliography{main}

\clearpage
\appendix
\label{sec:appendix}
\onecolumn

\section{Experimental Setup}
\subsection{Datasets}
\label{sec:appendix:datasets}
To ensure a rigorous and fair evaluation, we chose 31 benchmark datasets from the UCR Time Series Classification (TSC) Archive, a widely adopted repository for benchmarking TSC methods. The UCR archive provides a broad range of domains such as medical recordings, human activity and spectrography, making it an ideal testbed for evaluating generalizability. Our selection spans a broad spectrum of time-series characteristics, including varying sequence lengths, class counts, sample sizes, and both balanced and imbalanced class distributions. We prioritized datasets that could be meaningfully represented using chart-based visual encodings, making them well-suited for our investigation into visual and multimodal time-series representations. This curated subset enables us to assess generalization across diverse real-world signal types while maintaining consistency in experimental design. A detailed summary of dataset characteristics is provided in Table \ref{tab:dataset_stats}.

\begin{table}[H]
\centering
\normalsize
\caption{\textbf{Dataset statistics across different time-series UCR archive benchmark datasets.} Each dataset is described by its domain, data type, time-series length, number of train/test samples and number of classes.}
\begin{tabular}{r|l|l|l|c|c|c|c|c}
\toprule
\textbf{ID} & \textbf{Name} & \textbf{Domain} & \textbf{Type} & \textbf{\# Timesteps} & \textbf{\# Train} & \textbf{\# Test} & \textbf{\# Total} & \textbf{\# Class} \\
\midrule
1  & Adiac                      & Biology         & Image  & 176  & 390  & 391   & 781   & 37   \\
2  & Arrowhead                  & Anthropology        & Image         & 251  & 36   & 175   & 211   & 3    \\
3  & Beef                       & Food Science  & Spectro         & 470  & 30   & 30    & 60    & 5    \\
4  & Beetlefly                  & Biology       & Image         & 512  & 20   & 20    & 40    & 2   \\
5  & ChlorineConcentration      & Chemistry     & Simulation     & 166  & 467  & 3840  & 4307  & 3    \\
6  & Computers                  & Energy      & Device         & 720  & 250  & 250   & 500   & 2    \\
7  & CricketX                   & Motion        & Human Activity         & 300  & 390  & 390   & 780   & 12   \\
8  & CricketY                   & Motion        & Human Activity         & 300  & 390  & 390   & 780   & 12   \\
9  & CricketZ                   & Motion        & Human Activity         & 300  & 390  & 390   & 780   & 12   \\
10 & Crop                       & Agriculture             & Image       & 46   & 7200 & 16800 & 24000 & 24  \\
11 & ECG5000                    & Healthcare    & Sensor      & 140  & 500  & 4500  & 5000  & 5   \\
12 & Earthquakes                & Geophysics    & Sensor      & 512  & 322  & 139   & 461   & 2    \\
13 & FaceAll                    & Biometrics             & Image       & 131  & 560  & 1690  & 2250  & 14   \\
14 & FacesUCR                   & Biometrics             & Image       & 131  & 200  & 2050  & 2256  & 14   \\
15 & FordB                      & Automotive    & Sensor      & 500  & 3636 & 810   & 4446  & 2    \\
16 & GunPoint                   & Gesture       & Human Activity         & 150  & 50   & 150   & 200   & 2    \\
17 & Ham                        & Food Science  & Spectro     & 431  & 100  & 105   & 205   & 2    \\
18 & Herring                    & Biology       & Image       & 512  & 64   & 64    & 128   & 2    \\
19 & InsectWingbeatSound        & Biology       & Audio       & 256  & 220  & 1980  & 2200  & 10   \\
20 & ItalyPowerDemand           & Energy        & Sensor      & 24   & 67   & 1029  & 1096  & 2    \\
21 & Lightning2                 & Weather       & Sensor      & 637  & 60   & 61    & 121   & 2    \\
22 & PhalangesOutlinesCorrect   & Healthcare             & Image       & 80   & 1800 & 858   & 2658  & 2    \\
23 & RefrigerationDevices       & Energy    & Device      & 720  & 375  & 375   & 750   & 3    \\
24 & SonyAIBORobotSurface1      & Robotics      & Sensor      & 70   & 20   & 601   & 621   & 2    \\
25 & Strawberry                 & Food Science   & Spectro     & 235  & 613  & 370   & 983   & 2    \\
26 & ToeSegmentation1           & Motion       & Motion      & 277  & 40   & 228   & 268   & 2    \\
27 & ToeSegmentation2           & Motion       & Motion      & 343  & 36   & 130   & 166   & 2    \\
28 & Wafer                      & Manufacturing & Sensor      & 152  & 1000 & 6164  & 7164  & 2    \\
29 & Wine                       & Chemistry     & Spectro     & 234  & 57   & 54    & 111   & 2   \\
30 & WordSynonyms               & NLP           & Image       & 270  & 267  & 638   & 1162  & 25   \\
31 & Yoga                       & Motion & Image       & 426  & 300  & 3000  & 3300  & 2    \\
\bottomrule
\end{tabular}
\label{tab:dataset_stats}
\end{table}

While the UCR Archive provides a default train-test split for each dataset, we further split the test set into stratified validation and test partitions (80/20 split) using a fixed random seed. This protocol enables fair model selection via early stopping and hyperparameter tuning, without leaking information to the evaluation set.  

\subsection{Baselines}
\label{sec:appendix:baselines}
For baseline comparisons, we use published results reported in the original papers, as retraining all SOTA models under our protocol was computationally intensive. Results for GPT4TS, TimesNet, and SoftShape are taken directly from the TimesNet paper~\cite{TimesNet}. Nonetheless, our consistent evaluation pipeline allows for systemic and reproducible analysis of visual and multimodal representations under a unified benchmark. 

\subsubsection*{Ensemble and Tree-based Models}
\begin{itemize}
    \item \textbf{COTE}~\cite{bagnall2017great}: An ensemble combining multiple classifiers across diverse time-series representations (shape-based, spectral, autocorrelation) via weighted voting.
    \item \textbf{HC2 (HIVE-COTE 2.0)}~\cite{middlehurst2021hive}: A hierarchical meta-ensemble of specialized time-series classifiers, integrating interval-based, dictionary-based, and shapelet classifiers for high-accuracy time-series classification.
    \item \textbf{CHIEF (TS-CHIEF)}~\cite{shifaz2020ts}: A heterogeneous decision forest building trees using shape-based, spectral, and interval features for scalable and accurate classification.
\end{itemize}

\subsubsection*{Convolutional and Hybrid Deep Models}
\begin{itemize}
    \item \textbf{OS-CNN}~\cite{tang2020omni}: A CNN architecture that captures temporal features at multiple scales using parallel convolutional blocks with varying receptive fields.
    \item \textbf{FCN-LSTM}~\cite{karim2017lstm}: A hybrid architecture combining convolutional layers for local pattern extraction with LSTM modules to model long-term dependencies.
    \item \textbf{IncTime (InceptionTime)}~\cite{ismail2020inceptiontime}: A deep CNN leveraging Inception modules with variable kernel sizes for multi-scale time-series feature learning.
    \item \textbf{ROCKET}~\cite{dempster2020rocket}: A lightweight approach applying thousands of random convolutional kernels, followed by a linear classifier on their aggregated responses.
\end{itemize}

\subsubsection*{Transformer and Foundation Models}
\begin{itemize}
    \item \textbf{TimesNet}~\cite{TimesNet}: Transforms time series into 2D tensors and applies temporal spectral blocks to capture multi-period patterns efficiently.
    \item \textbf{GPT4TS}~\cite{zhou2023one}: A large pre-trained foundation model adapting GPT-style Transformers for time-series tasks through time-aware tokenization and pre-training
\end{itemize}

\subsubsection*{Shapelet-based Models}
\begin{itemize}
    \item \textbf{SoftShape}~\cite{liu2025learning}: A differentiable shapelet-based method that learns soft shape representations for interpretable and accurate time-series classification.
\end{itemize}

\section{More Details on Time-Series Chart Image Creation}
\label{sec:appendix:chart_image_creation}
To convert raw univariate time-series data into visual formats suitable for CNN-based processing, we generate chart-based image representations using matplotlib. We control two key image-generation factors:
\begin{itemize}
     \item \textbf{Color Mode}: Charts are rendered in black-and-white or a fixed color (e.g., blue) to isolate the effect of color cues on model learning.
     \item \textbf{Label Inclusion}:  Controls whether axis labels, title, bounding box, and tick marks are present. When disabled, charts contain only the plotted signal, reducing visual noise.
\end{itemize}

\begin{figure}[H]
  \centering
  \caption{\textbf{Examples of chart-based visual representations across four chart types (line, area, bar, scatter) and four visual settings (color/monochrome × label/no-label). Visuals shown are generated from the ItalyPowerDemand dataset for illustration. }}
  \label{fig:chart_types}

  \newcommand{\cellheight}{3.6cm}

  \setlength{\tabcolsep}{4pt}
  \renewcommand{\arraystretch}{1.5}

  \begin{tabular}{>{\centering\arraybackslash}m{2cm} | 
                  >{\centering\arraybackslash}m{\cellheight} 
                  >{\centering\arraybackslash}m{\cellheight} 
                  >{\centering\arraybackslash}m{\cellheight} 
                  >{\centering\arraybackslash}m{\cellheight}}
    \toprule
    \textbf{Chart Type} & \textbf{Monochrome w/o Label} & \textbf{Monochrome w/ Label} & \textbf{Color w/o Label} & \textbf{Color w/ Label} \\
    \midrule

    \textbf{Line} &
    \includegraphics[width=\linewidth]{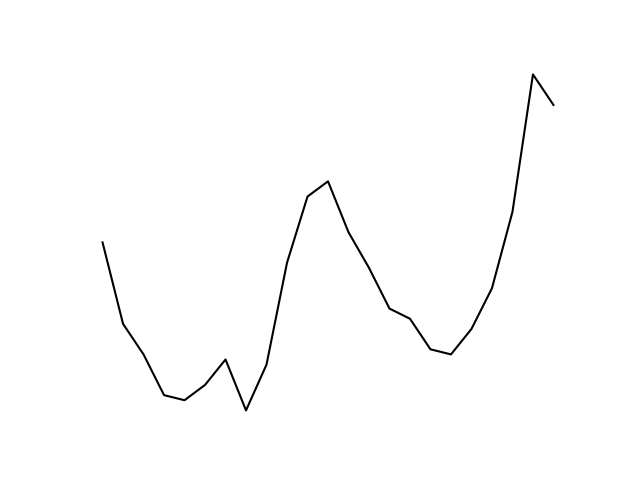} &
    \includegraphics[width=\linewidth]{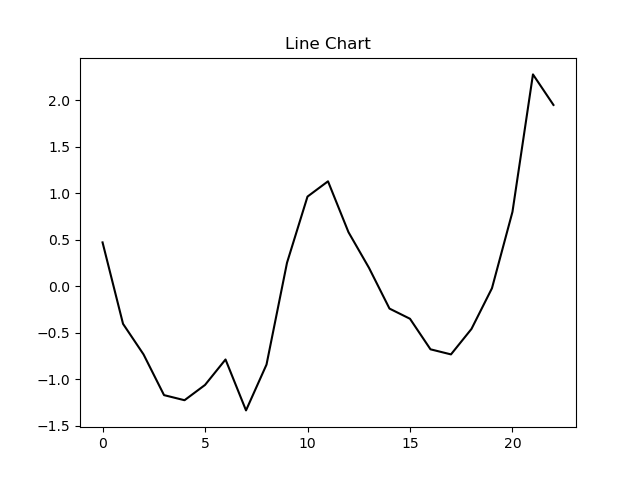} &
    \includegraphics[width=\linewidth]{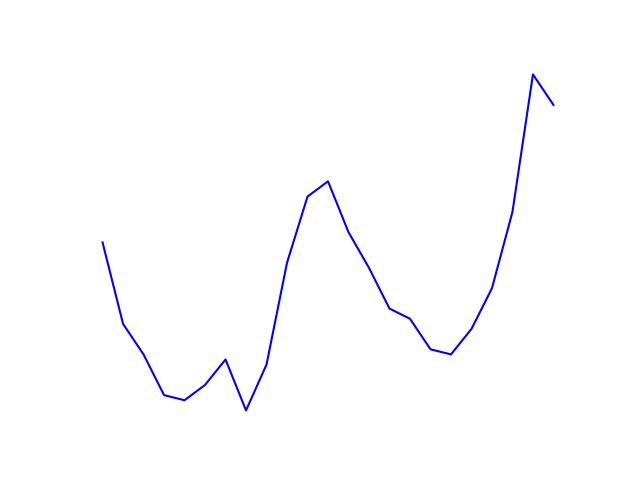} &
    \includegraphics[width=\linewidth]{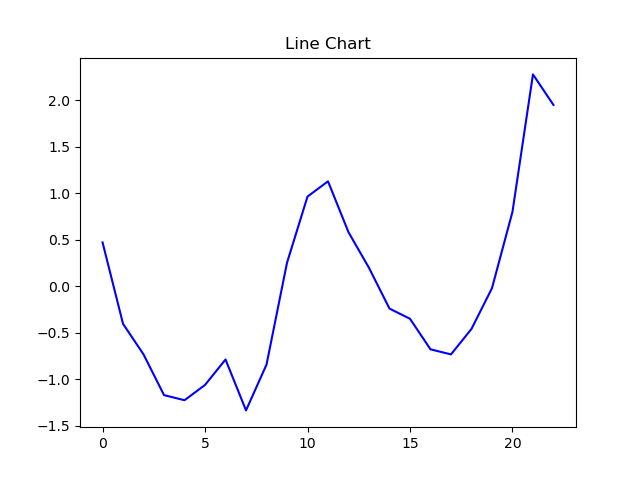} \\
    \midrule

    \textbf{Area} &
    \includegraphics[width=\linewidth]{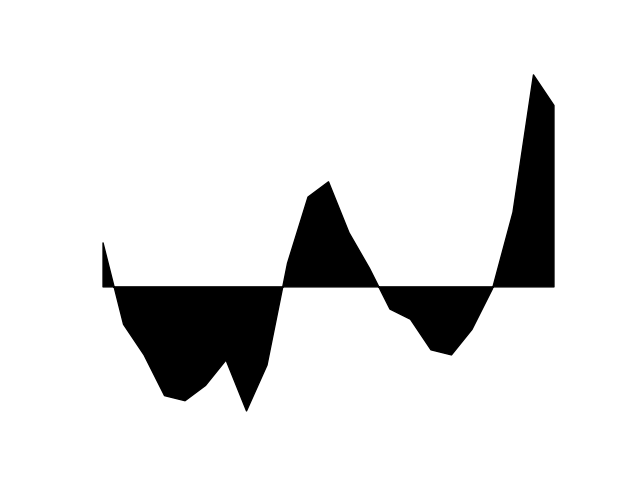} &
    \includegraphics[width=\linewidth]{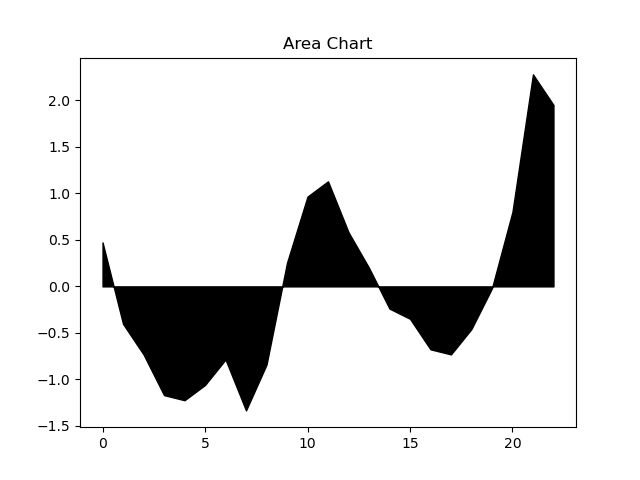} &
    \includegraphics[width=\linewidth]{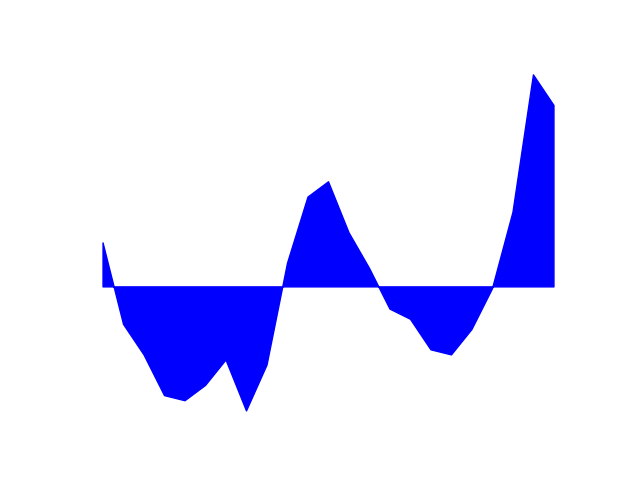} &
    \includegraphics[width=\linewidth]{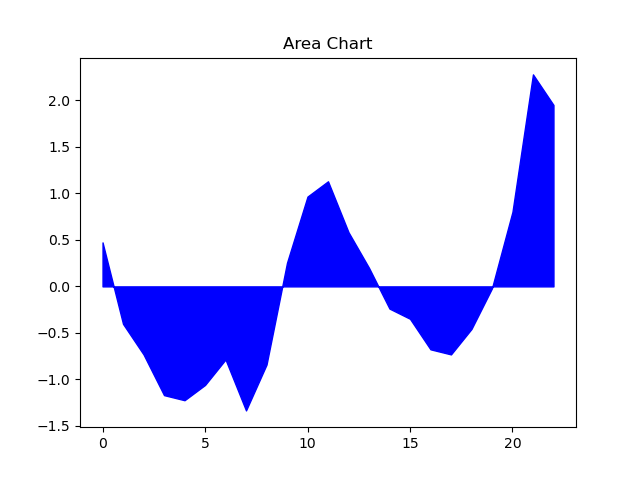} \\
    \midrule

    \textbf{Bar} &
    \includegraphics[width=\linewidth]{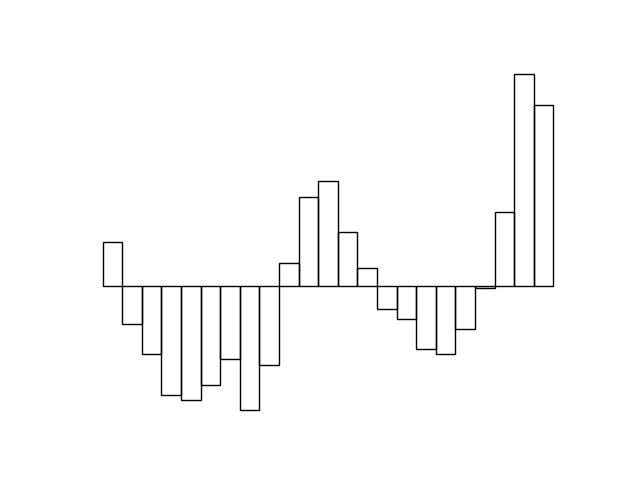} &
    \includegraphics[width=\linewidth]{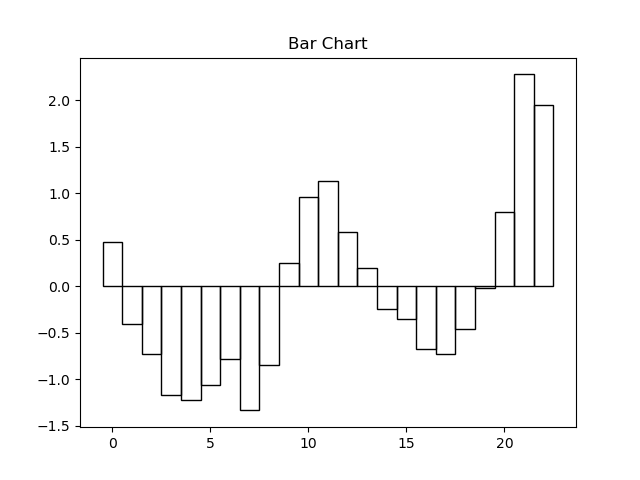} &
    \includegraphics[width=\linewidth]{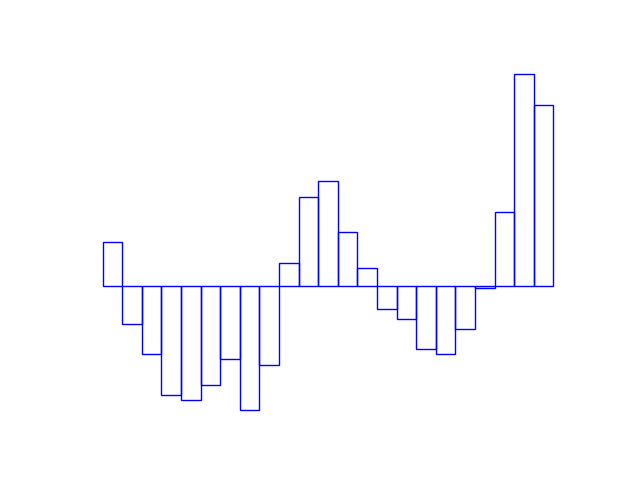} &
    \includegraphics[width=\linewidth]{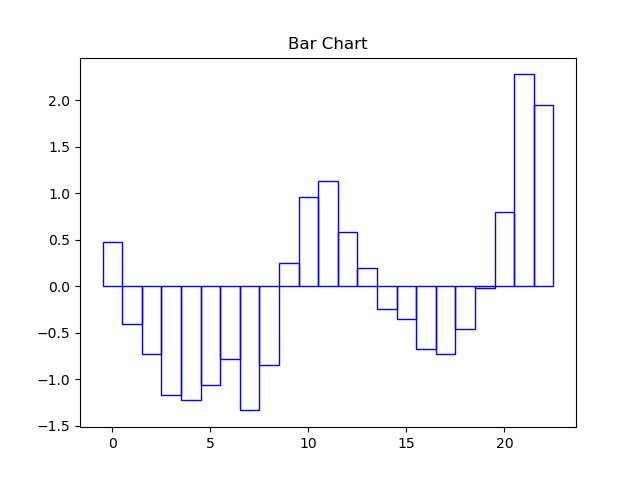} \\
    \midrule

    \textbf{Scatter} &
    \includegraphics[width=\linewidth]{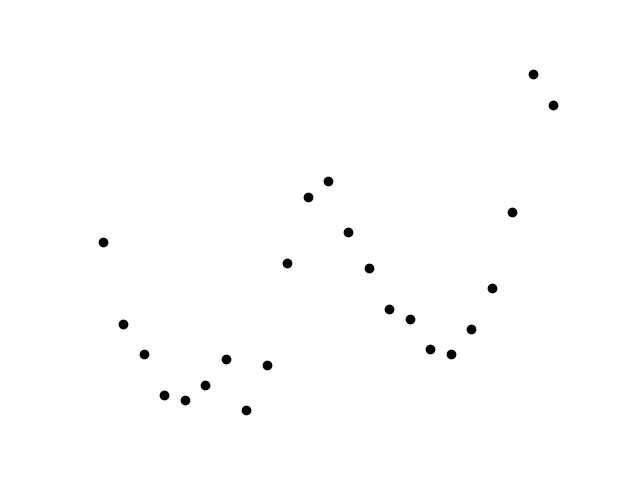} &
    \includegraphics[width=\linewidth]{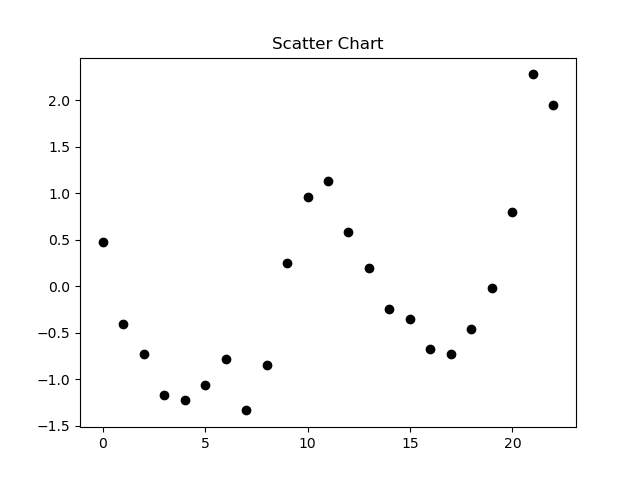} &
    \includegraphics[width=\linewidth]{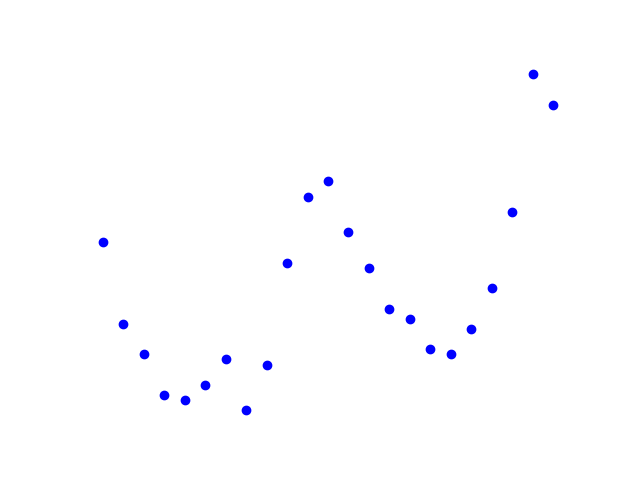} &
    \includegraphics[width=\linewidth]{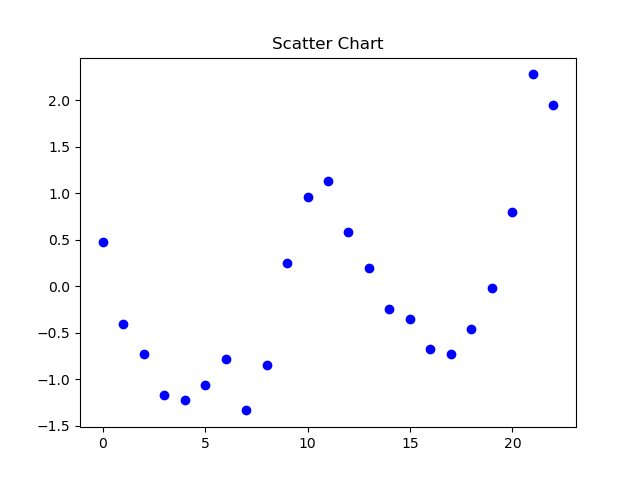} \\
    \bottomrule
  \end{tabular}
\end{figure}

All charts are generated on a clean white canvas with axis ticks, gridlines, and margins removed to eliminate visual noise and focus the model’s attention on shape-based cues. Below, we show the core function for creating an area chart from a univariate time-series input. The chart generation code is simple and lightweight, requiring only a few lines of code per chart type using matplotlib. Similar logic is applied for other chart types by switching the Matplotlib function used (e.g., \texttt{create\_bar\_chart}, \texttt{create\_line\_chart}, \texttt{create\_scatter\_chart}).
To ensure experimental consistency, we maintain a uniform chart generation pipeline across all datasets, applying the same procedure for each chart type and visual setting. This controlled setup allows us to isolate the effect of chart-based representations without confounding factors introduced by dataset-specific customization. The simplicity and modularity of the charting code (e.g., the area chart function below) enable scalable and reproducible chart generation for large-scale experiments. Figure \ref{fig:chart_types} shows example chart images from the same signal in ItalyPowerDemand dataset, spanning all 16 combinations of chart settings.

\lstdefinestyle{custompy}{
  language=Python,
  basicstyle=\ttfamily\footnotesize,
  keywordstyle=\color{green!50!black}\bfseries,
  stringstyle=\color{red!60!black},
  commentstyle=\color{teal}\itshape,
  identifierstyle=\color{black},
  showstringspaces=false,
  breaklines=true,
  frame=lines,
  tabsize=4,
  columns=fullflexible,
  keepspaces=true,
  captionpos=b,
  numbers=left,
  numberstyle=\tiny\color{gray},
}

\begin{lstlisting}[style=custompy, caption={Area chart generation using Matplotlib.}, label={lst:area_chart}]
def create_area_chart(ts, chart_path, color_mode, label_mode):
    plt.figure()
    plt.fill_between(range(len(ts)), ts,
                     color='blue' if color_mode == 'color' else 'black')
    if label_mode == 'with_label':
        plt.title('Area Chart')
    else:
        plt.axis('off')
    plt.savefig(chart_path, bbox_inches='tight', pad_inches=0)
    plt.close()
\end{lstlisting}

\section{Detailed Statistical Analysis}
\label{sec:appendix:statistical_analysis}
Tables \ref{tab:vtbench_class_ranks}–\ref{tab:type_rank_table} present a detailed breakdown of classifier performance across UCR datasets under different dataset characteristics. The results summarize average ranks (lower is better) and the number of first-place wins for each method when grouped by task type, training set size, sequence length, and data type. For reference, results for GPT4TS, TimesNet, and SoftShape are reported from the TimesNet paper~\cite{TimesNet}.

\begin{table}[H]
\centering
\normalsize
\caption{Average rank of classifiers across UCR datasets split by task type. Each cell shows the average rank, with the number of 1st-place wins in parentheses. Lower is better.}
\label{tab:vtbench_class_ranks}
\begin{tabular}{l|c|c|c|c}
\toprule
\textbf{Model} & \textbf{2 (N=17)} & \textbf{3-5 (N=5)} & \textbf{6-10 (N=1)} & \textbf{11+ (N=8)} \\
\midrule
COTE & 6.65 (0) & 6.40 (0) & 8.00 (0) & 6.43 (0) \\
HC2 & 5.18 (2) & 5.20 (0) & 4.00 (0) & 5.38 (0) \\
OS-CNN & 7.88 (0) & 7.80 (0) & 9.00 (0) & 6.00 (0) \\
FCN-LSTM & 4.47 (0) & 5.20 (0) & 6.00 (0) & 7.71 (0) \\
InceptionTime & 5.29 (1) & 6.40 (0) & 11.00 (0) & 4.00 (0) \\
ROCKET & 6.29 (0) & 7.40 (0) & 5.00 (0) & 5.62 (0) \\
TS-CHIEF & 6.41 (0) & 7.80 (0) & 10.00 (0) & 6.38 (0) \\
TimesNet & 5.53 (4) & 5.80 (0) & 3.00 (0) & 6.25 (0) \\
GPT4TS & 5.18 (3) & 3.60 (1) & 2.00 (0) & 6.00 (0) \\
SoftShape & 2.47 (6) & 1.60 (3) & 1.00 (1) & 1.00 (8) \\
\textbf{VTBench (Ours)} & \textbf{8.82 (0)} & \textbf{7.80 (0)} & 7.00 (0) & 10.00 (0) \\
\bottomrule
\end{tabular}
\end{table}

\begin{table}[H]
\centering
\normalsize
\caption{Average accuracy rank of classifiers grouped by training set size across UCR datasets. Each cell shows the average rank with number of 1st-place wins in parentheses.}
\label{tab:vtbench_train_size_ranks}
\begin{tabular}{l|c|c|c|c}
\toprule
\textbf{Model} & \textbf{<100 (11)} & \textbf{100–500 (14)} & \textbf{501–2000 (4)} & \textbf{>2000 (2)} \\
\midrule
COTE & 6.65 (0) & 6.40 (0) & 8.00 (0) & 6.43 (0) \\
HC2 & 5.18 (2) & 5.20 (0) & 4.00 (0) & 5.38 (0) \\
OS-CNN & 7.88 (0) & 7.80 (0) & 9.00 (0) & 6.00 (0) \\
FCN-LSTM & 4.47 (0) & 5.20 (0) & 6.00 (0) & 7.71 (0) \\
InceptionTime & 5.29 (1) & 6.40 (0) & 11.00 (0) & 4.00 (0) \\
ROCKET & 6.29 (0) & 7.40 (0) & 5.00 (0) & 5.62 (0) \\
TS-CHIEF & 6.41 (0) & 7.80 (0) & 10.00 (0) & 6.38 (0) \\
TimesNet & 5.53 (4) & 5.80 (0) & 3.00 (0) & 6.25 (0) \\
GPT4TS & 5.18 (3) & 3.60 (1) & 2.00 (0) & 6.00 (0) \\
SoftShape & 2.47 (6) & 1.60 (3) & 1.00 (1) & 1.00 (8) \\
\textbf{VTBench (Ours)} & \textbf{8.82 (0)} & \textbf{7.80 (0)} & \textbf{7.00 (0)} & \textbf{10.00 (0)} \\
\bottomrule
\end{tabular}
\end{table}

\begin{table}[H]
\centering
\normalsize
\caption{Average rank of classifiers across UCR datasets grouped by time-series length. Lower is better. Each cell shows average rank with number of 1st-place wins in parentheses.}
\label{tab:ts_length_ranks}
\begin{tabular}{l|c|c|c}
\toprule
\textbf{Model} & \textbf{<200 (11)} & \textbf{200–400 (10)} & \textbf{>400 (10)} \\
\midrule
COTE & 6.40 (0) & 7.10 (0) & 6.30 (0) \\
HC2 & 6.27 (0) & 4.10 (1) & 5.10 (0) \\
OS-CNN & 7.45 (0) & 6.40 (0) & 8.40 (0) \\
FCN-LSTM & 5.90 (0) & 5.60 (0) & 4.70 (0) \\
InceptionTime & 4.91 (0) & 4.70 (0) & 6.40 (1) \\
ROCKET & 6.00 (0) & 5.40 (0) & 7.40 (0) \\
TS-CHIEF & 6.91 (0) & 5.50 (0) & 7.80 (0) \\
TimesNet & 3.91 (3) & 8.10 (0) & 5.20 (1) \\
GPT4TS & 3.64 (0) & 8.00 (0) & 3.60 (4) \\
SoftShape & 1.91 (6) & 1.50 (9) & 2.30 (4) \\
\textbf{VTBench (Ours)} & \textbf{9.18 (0)} & \textbf{9.40 (0)} & \textbf{8.10 (0)} \\
\bottomrule
\end{tabular}
\end{table}

\begin{table}[H]
\centering
\normalsize
\caption{Average rank of classifiers grouped by data type across UCR datasets. Each cell shows the average rank with number of 1st-place wins in parentheses. Lower is better.}
\label{tab:type_rank_table}
\begin{tabular}{l|c|c|c|c|c|c|c}
\toprule
\textbf{Model} & \textbf{Image} & \textbf{Sensor} & \textbf{Spectro} & \textbf{Simulation} & \textbf{Device} & \textbf{Human Activity} & \textbf{Audio} \\
\midrule
COTE & 6.7 (0) & 5.4 (0) & 7.0 (0) & 10.0 (0) & 6.0 (0) & 7.2 (0) & 8.0 (0) \\
HC2 & 5.6 (0) & 6.0 (0) & 4.8 (1) & 9.0 (0) & 2.5 (0) & 4.2 (0) & 4.0 (0) \\
OS-CNN & 7.9 (0) & 8.1 (0) & 8.0 (0) & 6.0 (0) & 10.0 (0) & 3.8 (0) & 9.0 (0) \\
FCN-LSTM & 5.8 (0) & 5.3 (0) & 5.2 (0) & 7.0 (0) & 5.0 (0) & 5.8 (0) & 6.0 (0) \\
InceptionTime & 5.7 (0) & 6.3 (0) & 6.8 (0) & 5.0 (0) & 2.0 (1) & 2.2 (0) & 11.0 (0) \\
ROCKET & 6.6 (0) & 7.4 (0) & 5.5 (0) & 8.0 (0) & 5.5 (0) & 3.5 (0) & 5.0 (0) \\
TS-CHIEF & 6.9 (0) & 7.0 (0) & 7.8 (0) & 11.0 (0) & 7.0 (0) & 4.8 (0) & 10.0 (0) \\
TimesNet & 5.0 (1) & 3.3 (3) & 6.8 (0) & 1.0 (0) & 7.5 (0) & 9.5 (0) & 3.0 (0) \\
GPT4TS & 4.6 (1) & 2.6 (1) & 5.0 (2) & 1.0 (0) & 6.0 (0) & 9.5 (0) & 2.0 (0) \\
SoftShape & 1.2 (8) & 1.9 (2) & 3.2 (1) & 1.0 (0) & 3.0 (1) & 2.8 (3) & 1.0 (1) \\
\textbf{VTBench (Ours)} & \textbf{8.7 (0)} & \textbf{9.9 (0)} & \textbf{5.2 (0)} & \textbf{4.0 (0)} & \textbf{11.0 (0)} & \textbf{11.0 (0)} & \textbf{7.0 (0)} \\
\bottomrule
\end{tabular}
\end{table}

\begin{figure}[H]
  \centering
  \includegraphics[width=0.75\linewidth]{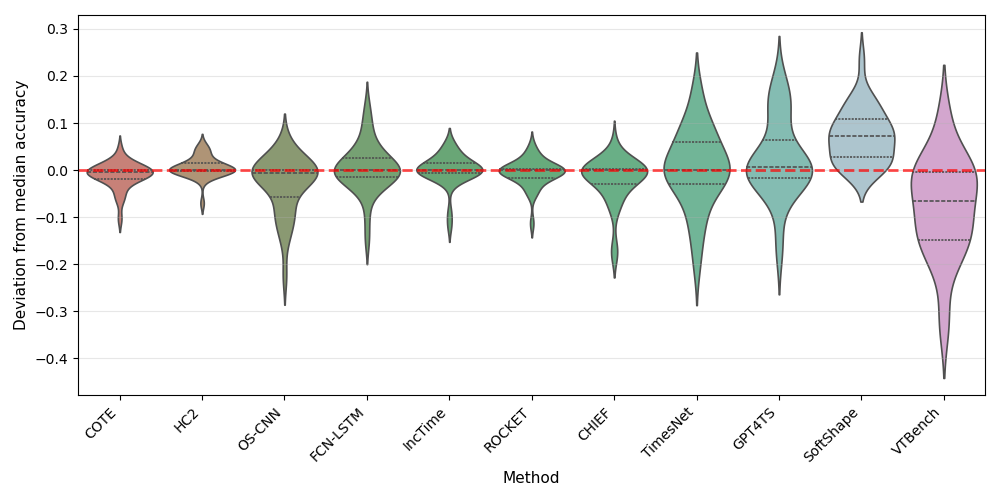}
  \caption{Violin plots showing the deviation of each classifier's accuracy from the dataset-wise median performance across 31 UCR datasets. Each violin represents the distribution of deviation values for a classifier, highlighting both its typical and outlier behavior relative to other methods. A horizontal dashed line indicates zero deviation from the median.}
  \label{fig:charts_violinplot}
\end{figure}
Figure \ref{fig:charts_violinplot} presents the distribution of accuracy deviations from the dataset-wise median for each method across 31 UCR datasets. For each dataset, we compute the median accuracy across all baseline models (excluding VTBench), then plot each method’s deviation from this value. Positive values indicate performance above the median, while negative values denote below-median outcomes. This analysis offers a normalized view of relative performance across datasets of varying difficulty. It reveals that VTBench exhibits a distinctly different performance profile compared to traditional time-series classifiers. While most baseline methods (COTE through CHIEF) demonstrate relatively narrow distributions centered near zero deviation, indicating consistent performance around the dataset median, VTBench shows a much wider distribution with a notable negative skew. This broader distribution suggests that VTBench's performance is more variable across datasets, with a tendency to underperform relative to the median on a substantial number of datasets (as evidenced by the extended lower tail reaching approximately -0.4 deviation). However, the presence of positive deviations in the upper portion of the distribution indicates that VTBench can achieve above-median performance on certain datasets. Interestingly, the more recent deep learning methods (TimesNet, GPT4TS, SoftShape) show progressively wider distributions and higher variance, with SoftShape and GPT4TS displaying more positive skew, suggesting these methods can achieve exceptional performance on some datasets while potentially struggling on others. VTBench's performance pattern aligns with this trend of increased variability in modern approaches, reflecting the inherent trade-offs between specialized high performance on certain data types versus consistent baseline performance across all domain

\section{Preliminary Study}
\label{sec:appendix:preliminary_study}

\begin{figure}[!t]
  \centering
  \includegraphics[width=0.6\columnwidth]{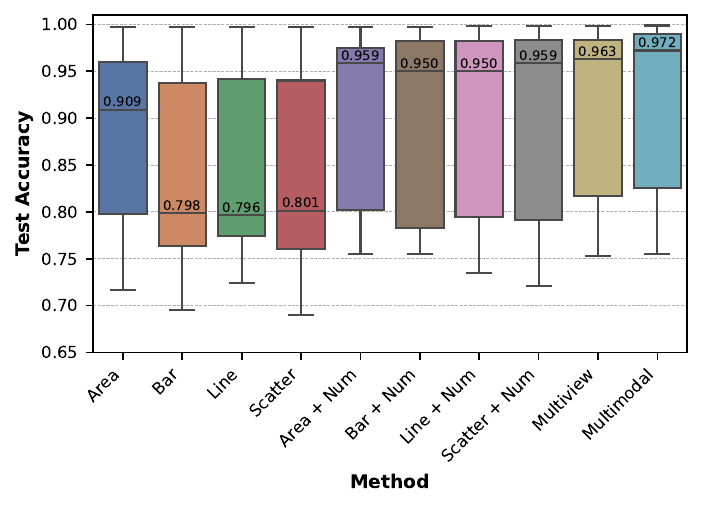}
  \caption{Exploratory results on 9 UCR datasets using fixed 640×480 chart resolution. Each boxplot summarizes the distribution of the best-performing unimodal chart type (line, bar, area, or scatter), the strongest multimodal variant (single chart + numerical input), the best multiview fusion of all chart types, and the top-performing full multimodal setup (multiview + numerical). Best scores are selected across all tested model-level configurations. These early findings show that combining multiple visual encodings and numerical input consistently improves accuracy, motivating the full-scale VTBench benchmark.}
  \label{fig:charts_boxplot}
\end{figure}
Before committing to a full-scale benchmark, we conducted a controlled exploratory study on 9 UCR datasets selected to span diverse characteristics relevant to chart-based modeling, including sequence length, training size, and domain. This ensured early findings reflected realistic variation in time-series properties. Eight datasets were naturally binary classification tasks, while ECG5000, originally five-class, was converted to binary by grouping its four minority classes to reduce task complexity while preserving meaningful signal separation (normal vs.abnormal patterns). Images were rendered at the default 640×480 resolution and internally downscaled to 64×64 by the CNN, serving as a stress test for robustness under non-optimized visualization settings. Results (Figure\ref{fig:charts_boxplot}) showed that (i) chart-only models had high variance across datasets, (ii) adding numerical input consistently improved accuracy, and (iii) combining multiple chart modalities with numerical input delivered the highest and most stable performance. These findings (details in Appendix \ref{sec:appendix}) provided the first empirical validation of VTBench’s hypothesis, motivating a broader benchmark with dataset-aware resizing and expanded coverage.
\begin{table}[H]
\centering
\small
\caption{\textbf{Exploratory feasibility results on 9 representative UCR datasets using fixed 640×480 chart resolution. The table reports: (i) the highest test accuracy achieved by any unimodal chart type (line, bar, area, scatter), (ii) the best multimodal variant combining a single chart type with numerical input, and (iii) the best-performing multiview and full multimodal configurations, selected from all tested model-level settings (OS-CNN vs. Transformer encoder and concatenation vs. dynamic weighted fusion). These exploratory findings, obtained under a non-optimized fixed resolution, provide early evidence that fusing multiple chart encodings and numerical input can consistently outperform single-chart baselines, motivating the full-scale VTBench benchmark.}}
\label{tab:vtbench_chart_comparison}
\begin{tabular}{c|c|c|c|c|c|c|c|c|c|c}
\toprule
\textbf{Dataset} & \textbf{Area} & \textbf{Bar} & \textbf{Line} & \textbf{Scatter} & \textbf{Area + Num} & \textbf{Bar + Num} & \textbf{Line + Num} & \textbf{Scatter + Num} & \textbf{Multiview} & \textbf{Multimodal} \\
\midrule
Strawberry & 0.9406 & 0.9378 & 0.9369 & 0.9401 & 0.9594 & 0.9500 & 0.9500 & 0.9719 & 0.9628 & \textbf{0.9764} \\
Yoga & 0.8071 & 0.7985 & 0.7962 & 0.8012 & 0.8165 & 0.8315 & 0.8433 & 0.8406 & 0.8171 & \textbf{0.8258} \\
FordB & 0.7171 & 0.7153 & 0.7242 & 0.7139 & 0.7592 & 0.7737 & 0.7724 & 0.7211 & 0.7531 & \textbf{0.7546} \\
ECG5000 (Binary) & 0.9852 & 0.9434 & 0.9844 & 0.9837 & 0.9817 & 0.9824 & 0.9822 & 0.9855 & 0.9883 & \textbf{0.99} \\
GunPoint & 0.9090 & 0.6951 & 0.7701 & 0.6904 & 0.9750 & 0.9833 & 0.9917 & 0.9833 & 0.9833 & \textbf{0.9917} \\
Lightning2 & 0.7715 & 0.7792 & 0.7852 & 0.7754 & 0.7551 & 0.7551 & 0.7347 & 0.7355 & \textbf{0.8571} & \textbf{0.8571} \\
ItalyPowerDemand & 0.9594 & 0.8816 & 0.9420 & 0.8603 & 0.9587 & 0.9733 & 0.9697 & 0.9587 & 0.9697 & \textbf{0.9721} \\
PhalangesOutlinesCorrect & 0.7976 & 0.7638 & 0.7747 & 0.7601 & 0.8020 & 0.7831 & 0.7948 & 0.7918 & 0.7977 & \textbf{0.8137} \\
Wafer & 0.9975 & 0.9972 & 0.9973 & 0.9971 & 0.9972 & 0.9968 & 0.9982 & 0.9984 & 0.9986 & \textbf{0.9988} \\
\bottomrule
\end{tabular}
\end{table}

\begin{table}[H]
\centering
\normalsize
\caption{\textbf{Preliminary comparison of CNN architectures (ShallowCNN vs. DeepCNN) across datasets at a fixed resolution of 640×480.} 
Results are reported for the best-performing chart type per dataset (Area or Line). Best architecture and selected metric scores per dataset are highlighted in bold.}

\label{tab:chart_model_comparison}
\begin{tabular}{cccccc}
\toprule
\textbf{Dataset} & \textbf{Chart Type} & \textbf{Architecture} & \textbf{Accuracy} & \textbf{F1 Score} & \textbf{AUC} \\
\midrule

\multirow{2}{*}{Strawberry} & \multirow{2}{*}{Area} & ShallowCNN & 0.9249 & 0.9407 & 0.9857 \\
                            &                        & \textbf{DeepCNN} & \textbf{0.9406} & \textbf{0.9536} & \textbf{0.9868} \\

\multirow{2}{*}{Yoga} & \multirow{2}{*}{Area} & \textbf{ShallowCNN} & \textbf{0.8071} & \textbf{0.8262} & \textbf{0.8916} \\
                      &                       & DeepCNN & 0.7962 & 0.8215 & 0.8858 \\

\multirow{2}{*}{FordB} & \multirow{2}{*}{Line} & \textbf{ShallowCNN} & \textbf{0.7242} & \textbf{0.6963} & \textbf{0.8519} \\
                       &                        & DeepCNN & 0.7148 & 0.6817 & 0.8388 \\

\multirow{2}{*}{ECG5000 (Binary)} & \multirow{2}{*}{Area} & \textbf{ShallowCNN} & \textbf{0.9852} & \textbf{0.9711} & \textbf{0.9989} \\
                       &                        & DeepCNN & 0.9833 & 0.9796 & 0.9986 \\

\multirow{2}{*}{GunPoint} & \multirow{2}{*}{Area} & \textbf{ShallowCNN} & \textbf{0.9090} & \textbf{0.9251} & \textbf{0.9899} \\
                       &                        & DeepCNN & 0.6619 & 0.7534 & 0.9711 \\

\multirow{2}{*}{Lightning2} & \multirow{2}{*}{Line} & \textbf{ShallowCNN} & \textbf{0.7852} & \textbf{0.8478} & \textbf{0.9777} \\
                       &                        & DeepCNN & 0.7204 & 0.8081 & 0.9596 \\

\multirow{2}{*}{ItalyPowerDemand} & \multirow{2}{*}{Area} & \textbf{ShallowCNN} & \textbf{0.9594} & \textbf{0.9589} & \textbf{0.9906} \\
                       &                        & DeepCNN & 0.9594 & 0.9589 & 0.9907 \\

\multirow{2}{*}{PhalangesOutlinesCorrect} & \multirow{2}{*}{Area} & \textbf{ShallowCNN} & \textbf{0.7976} & \textbf{0.8417} & \textbf{0.8622} \\
                       &                        & DeepCNN & 0.7593 & 0.8156 & 0.8409 \\

\multirow{2}{*}{Wafer} & \multirow{2}{*}{Area} & \textbf{ShallowCNN} & \textbf{0.9975} & \textbf{0.9986} & \textbf{0.9998} \\
                       &                        & DeepCNN & 0.9969 & 0.9982 & 0.9998 \\

\bottomrule
\end{tabular}

\end{table}

\textit{Note:} This preliminary analysis was conducted under early experimental settings, where all images were initially generated at 640×480 and downsampled internally to 64×64. ShallowCNN achieved higher accuracy on most datasets under this low-resolution setup. However, we did not evaluate ShallowCNN at the tuned resolution of 128×128 determined in later experiments. DeepCNN was selected for all main experiments due to its higher representational capacity and superior scalability at higher resolutions.

\clearpage
\section{Full Results}
\begin{table}[h]
\renewcommand{\arraystretch}{1.2}
\definecolor{medgreen}{RGB}{180, 240, 180}
\centering
\normalsize
\caption{Comparison of Best Single Chart vs. VTBench Multimodal results.
\textbf{Default} uses consistent chart settings across all branches, while \textbf{Tuned} displays the upper bound of performance if the best chart settings were used individually across all branches.}
\label{tab:charts_numerical_ablation}
\begin{tabular}{l| c | c | c | c | c | c}
\toprule
\multirow{2}{*}{\textbf{Dataset}} &
\multirow{2}{*}{\textbf{Transformer}} & 
\multirow{2}{*}{\textbf{Best Single}} &
\multirow{2}{*}{\textbf{Best Single}} &
\multicolumn{2}{c}{\textbf{VTBench Multimodal}} \\
\cmidrule(lr){5-6}
& & \textbf{Chart} & \textbf{Chart + Numerical} & \textbf{Default} & \textbf{Tuned} \\
\midrule
Strawberry &0.974& 0.976 & \cellcolor{medgreen}0.983 &\cellcolor{medgreen} 0.986 & 0.973 \\
Adiac & 0.786& 0.631 & \cellcolor{medgreen}0.788 & \cellcolor{medgreen}0.776 & 0.792 \\
ECG5000 & 0.927& 0.937 & \cellcolor{medgreen}0.941 & \cellcolor{medgreen}0.939 & 0.940 \\
Beetlefly & 0.869 & 0.925 & 0.875 & \cellcolor{medgreen}0.938 & 0.875 \\
Lightning2 & 0.689 & 0.846 & 0.750 & \cellcolor{medgreen}0.771 & \cellcolor{medgreen}0.792 \\
ItalyPowerDemand & 0.948 & 0.969 & \cellcolor{medgreen}0.972 & 0.967 & 0.968 \\
PhalangesOutlinesCorrect &0.801 & 0.817 & 0.802 & 0.816 & \cellcolor{medgreen}0.821 \\
Arrowhead & 0.697 & 0.791 & \cellcolor{medgreen}0.736 & 0.714 & \cellcolor{medgreen}0.736 \\
Earthquakes & 0.649 & 0.738 & 0.712 & 0.721 & \cellcolor{medgreen}0.739 \\
Beef & 0.936 & 0.750 & \cellcolor{medgreen}0.875 & \cellcolor{medgreen}0.917 & \cellcolor{medgreen}0.833 \\
CricketZ & 0.529 & 0.755 & 0.673 & 0.676 & 0.676 \\
WordSynonyms & 0.567& 0.681 & 0.594 & 0.629 & 0.635 \\
CricketX & 0.492& 0.750 & 0.654 & 0.670 & 0.679 \\
CricketY & 0.548& 0.730 & 0.670 & 0.718 & 0.728 \\
Computers & 0.533& 0.704 & 0.625 & 0.605 & 0.630 \\
ChlorineConcentration & 0.909 & 0.752 & \cellcolor{medgreen}0.872 & \cellcolor{medgreen}0.868 & \cellcolor{medgreen}0.858 \\
FaceAll &0.872 & 0.821 & \cellcolor{medgreen}0.879 & \cellcolor{medgreen}0.876 & \cellcolor{medgreen}0.895 \\
FaceUCR &0.798 & 0.952 & 0.834 & 0.829 & 0.837 \\
Ham & 0.693& 0.703 & \cellcolor{medgreen}0.726 & \cellcolor{medgreen}0.714 & 0.679 \\
Herring & 0.694& 0.598 & \cellcolor{medgreen}0.804 & \cellcolor{medgreen}0.745 & \cellcolor{medgreen}0.686 \\
RefrigerationDevices &0.347 & 0.543 & 0.387 & 0.393 & 0.397 \\
SonyAIBORobotSurface1 & 0.737  & 0.831 & 0.796 & 0.798 & 0.783 \\
ToeSegmentation1 & 0.589 & 0.921 & 0.593 & 0.643 & 0.676 \\
ToeSegmentation2 &0.747& 0.938 & 0.808 & 0.750 & 0.760 \\
Wine & 0.741 & 0.751 & 0.721 & \cellcolor{medgreen}0.860 & 0.814 \\
GunPoint & 0.938 & 0.997 & 0.983 & 0.933 & 0.967 \\
Wafer & 0.995 & 0.997 & 0.996 & \cellcolor{medgreen}0.997 & 0.996 \\
Crop &0.769 & 0.776 & \cellcolor{medgreen}0.789 & \cellcolor{medgreen}0.789 & \cellcolor{medgreen}0.791\\
InsectWingBeatSound & 0.579 & 0.653 & 0.640 & 0.650 & 0.650 \\
Yoga & 0.823 & 0.876 & 0.832 & 0.807 & 0.828 \\
FordB & 0.708 & 0.782 & \cellcolor{medgreen}0.824 & \cellcolor{medgreen}0.809 & \cellcolor{medgreen}0.806 \\
\midrule
\textbf{Average} & 0.738 & \textbf{0.803} & \textbf{0.779} & \textbf{0.784} & \textbf{0.782} \\
\bottomrule
\end{tabular}
\end{table}

\clearpage
\section{Ablation Study}
\label{sec:appendix:ablation_study}

\definecolor{lightgreen}{RGB}{220, 255, 220}
\definecolor{medgreen}{RGB}{180, 240, 180}
\definecolor{darkgreen}{RGB}{140, 220, 140}
\begin{table}[H]
\centering
\setlength{\tabcolsep}{5pt}
\normalsize
\caption{\textbf{Impact of image size on classification accuracy.} Datasets are grouped by time-series length. Bold values indicate the best-performing configuration per row. Color intensity reflects relative accuracy per dataset.}
\label{tab:image_size_by_length}
\begin{tabular}{ll|c|c|c|c}
\toprule
\textbf{Length Group} & \textbf{Dataset} & \textbf{Length} & \textbf{64×64} & \textbf{128×128} & \textbf{256×256} \\
\midrule
\multirow{11}{*}{Short (<200)} 
& Crop             & 46 & \cellcolor{lightgreen}0.7221 & \cellcolor{medgreen}0.7233 & \cellcolor{darkgreen} \textbf{0.7293} \\
& SonyAIBORobotSurface1             & 70 & \cellcolor{lightgreen}0.6667 & \cellcolor{medgreen} 0.7426 & \cellcolor{darkgreen} \textbf{0.7696} \\
& ChlorineConcentration             & 166 & \cellcolor{lightgreen}0.6198 & \cellcolor{medgreen} 0.6751 & \cellcolor{darkgreen} 0.6838 \\
& Wafer            & 152 & \cellcolor{medgreen} 0.9966 & \cellcolor{darkgreen} \textbf{0.9967} & \cellcolor{lightgreen} 0.9956 \\
& GunPoint         & 52  & \cellcolor{lightgreen} 0.9644 & \cellcolor{medgreen} \textbf{0.9711} & \cellcolor{darkgreen} 0.9822 \\
& ECG5000          & 140 & \cellcolor{lightgreen} 0.9318 & \cellcolor{darkgreen} \textbf{0.9384} & \cellcolor{medgreen} 0.9367 \\
& Phalanges        & 80  & \cellcolor{medgreen} 0.7983 & \cellcolor{darkgreen} \textbf{0.8003} & \cellcolor{lightgreen} 0.7933 \\
& ItalyPowerDemand & 24  & \cellcolor{lightgreen} 0.9578 & \cellcolor{medgreen} 0.9604 & \cellcolor{darkgreen} \textbf{0.9615} \\
& Adiac & 176  & \cellcolor{lightgreen} 0.6292 & \cellcolor{darkgreen} \textbf{0.6419} & \cellcolor{medgreen} \textbf{0.6308} \\
& FaceAll & 131  & \cellcolor{lightgreen} 0.7919 & \cellcolor{medgreen} 0.8048 & \cellcolor{darkgreen} \textbf{0.9512} \\
& FacesUCR & 131  & \cellcolor{lightgreen} 0.8371 & \cellcolor{medgreen} 0.8694 & \cellcolor{darkgreen} \textbf{0.9005} \\
\midrule
\multirow{9}{*}{Medium (200–400)} 
& Strawberry       & 235 & \cellcolor{lightgreen} 0.9594 & \cellcolor{medgreen} 0.9648 & \cellcolor{darkgreen} \textbf{0.9649} \\
& ToeSegmentation1       & 277 & \cellcolor{darkgreen} \textbf{0.9836} & \cellcolor{medgreen} 0.8421 & \cellcolor{lightgreen} 0.7544 \\
& ToeSegmentation2       & 343 & \cellcolor{lightgreen} 0.8974 & \cellcolor{medgreen} 0.7718 & \cellcolor{darkgreen} \textbf{0.7538} \\
& CricketX       & 300 & \cellcolor{lightgreen} 0.6607 & \cellcolor{medgreen} 0.6683 & \cellcolor{darkgreen} 0.6948 \\
& CricketY       & 300 & \cellcolor{lightgreen} 0.6752 & \cellcolor{medgreen} 0.6940 & \cellcolor{darkgreen} \textbf{0.7068} \\
& CricketZ       & 300 & \cellcolor{lightgreen} 0.7085 & \cellcolor{darkgreen} \textbf{0.7367} & \cellcolor{medgreen} 0.7324 \\
& WordSynonyms       & 270 & \cellcolor{medgreen} 0.6081 & \cellcolor{medgreen} 0.6457 & \cellcolor{darkgreen} \textbf{0.6541} \\
& Arrowhead       & 251 & \cellcolor{darkgreen} \textbf{0.7580} & \cellcolor{medgreen} 0.7390 & \cellcolor{lightgreen} 0.7333 \\
& Wine       & 234 & \cellcolor{medgreen} 0.6728 & \cellcolor{darkgreen} \textbf{0.7160} & \cellcolor{lightgreen} 0.6481 \\

\midrule
\multirow{10}{*}{Long (>400)} 
& FordB            & 500 & \cellcolor{lightgreen} 0.7493 & \cellcolor{medgreen} 0.7884 & \cellcolor{darkgreen} \textbf{0.8004} \\
& Ham            & 431 & \cellcolor{lightgreen} 0.6603 & \cellcolor{darkgreen} \textbf{0.7047} & \cellcolor{medgreen} 0.6825 \\
& Beef            & 470 & \cellcolor{medgreen} 0.5667 & \cellcolor{darkgreen} \textbf{0.5888} & \cellcolor{lightgreen} 0.5444 \\
& Beetlefly            & 512 & \cellcolor{medgreen} 0.8000 & \cellcolor{medgreen}0.8333 & \cellcolor{darkgreen} \textbf{0.8667} \\
& Computers            & 720 & \cellcolor{lightgreen} 0.8570 & \cellcolor{darkgreen} \textbf{0.8746} & \cellcolor{medgreen} 0.8723 \\
& Earthquakes            & 512 & \cellcolor{lightgreen} 0.7505 & \cellcolor{darkgreen} \textbf{0.7649} & \cellcolor{medgreen} 0.7529 \\
& Herring            & 512 & \cellcolor{darkgreen} \textbf{0.5989} &\cellcolor{lightgreen} 0.5625 & \cellcolor{medgreen} 0.5677 \\
& RefrigerationDevices            & 720 & \cellcolor{lightgreen} 0.5688 & \cellcolor{medgreen} 0.6115 & \cellcolor{darkgreen} \textbf{0.6186} \\
& Yoga             & 426 & \cellcolor{lightgreen} 0.8203 & \cellcolor{medgreen} 0.8284 & \cellcolor{darkgreen} \textbf{0.8356} \\
& Lightning2       & 637 & \cellcolor{darkgreen} \textbf{0.8087} & \cellcolor{lightgreen} 0.7814 & \cellcolor{medgreen} 0.7486 \\
& InsectSound       & 600 & \cellcolor{lightgreen} 0.6683 & \cellcolor{medgreen} 0.6896 & \cellcolor{darkgreen} \textbf{0.7000} \\
\bottomrule
\end{tabular}
\end{table}

\begin{table*}[!t]
\centering
\scriptsize
\renewcommand{\arraystretch}{0.95}
\caption{\textbf{Test accuracy for Line and Area chart types under four rendering settings (image size = 128$\times$128).}}
\label{tab:line_area_settings_scaled}
\begin{tabular}{l|cccc|cccc}
\toprule
\multirow{2}{*}{\textbf{Dataset}} & 
\multicolumn{4}{c|}{\textbf{Line}} & 
\multicolumn{4}{c}{\textbf{Area}} \\
& Monochrome w Label & Monochrome w/o Label & Color w Label & Color w/o Label & Monochrome w Label & Monochrome w/o Label & Color w Label & Color w/o Label \\
\midrule
Adiac & 0.642 & 0.631 & 0.627 & 0.627 & 0.617 & 0.617 & 0.606 & 0.614 \\
Arrowhead & 0.577 & 0.520 & 0.547 & 0.415 & 0.739 & 0.644 & 0.722 & 0.585 \\
Beef & 0.444 & 0.578 & 0.422 & 0.367 & 0.522 & 0.400 & 0.456 & 0.411 \\
Beetlefly & 0.750 & 0.767 & 0.717 & 0.617 & 0.833 & 0.733 & 0.800 & 0.700 \\
ChlorineConcentration & 0.646 & 0.676 & 0.650 & 0.636 & 0.640 & 0.609 & 0.601 & 0.585 \\
Computers & 0.761 & 0.767 & 0.757 & 0.801 & 0.659 & 0.644 & 0.685 & 0.675 \\
CricketX & 0.668 & 0.678 & 0.643 & 0.666 & 0.648 & 0.657 & 0.625 & 0.659 \\
CricketY & 0.622 & 0.694 & 0.659 & 0.693 & 0.630 & 0.670 & 0.617 & 0.662 \\
CricketZ & 0.685 & 0.720 & 0.680 & 0.737 & 0.690 & 0.715 & 0.676 & 0.712 \\
Crop & 0.717 & 0.702 & 0.720 & 0.703 & 0.723 & 0.717 & 0.726 & 0.712 \\
ECG5000 & 0.931 & 0.938 & 0.934 & 0.932 & 0.929 & 0.936 & 0.929 & 0.933 \\
Earthquakes & 0.753 & 0.748 & 0.724 & 0.758 & 0.727 & 0.717 & 0.715 & 0.736 \\
FaceAll & 0.754 & 0.816 & 0.784 & 0.815 & 0.776 & 0.764 & 0.760 & 0.793 \\
FacesUCR & 0.847 & 0.865 & 0.823 & 0.856 & 0.854 & 0.866 & 0.833 & 0.869 \\
FordB & 0.735 & 0.783 & 0.735 & 0.788 & 0.739 & 0.778 & 0.740 & 0.786 \\
GunPoint & 0.964 & 0.927 & 0.940 & 0.971 & 0.927 & 0.918 & 0.938 & 0.931 \\
Ham & 0.629 & 0.698 & 0.673 & 0.644 & 0.603 & 0.667 & 0.603 & 0.632 \\
Herring & 0.500 & 0.563 & 0.479 & 0.536 & 0.453 & 0.500 & 0.474 & 0.469 \\
InsectWingbeatSound & 0.643 & 0.690 & 0.638 & 0.689 & 0.637 & 0.690 & 0.634 & 0.683 \\
ItalyPowerDemand & 0.952 & 0.956 & 0.941 & 0.927 & 0.948 & 0.954 & 0.949 & 0.960 \\
Lightning2 & 0.661 & 0.716 & 0.694 & 0.738 & 0.738 & 0.765 & 0.699 & 0.743 \\
PhalangesOutlinesCorrect & 0.785 & 0.771 & 0.800 & 0.775 & 0.786 & 0.787 & 0.786 & 0.787 \\
RefrigerationDevices & 0.564 & 0.552 & 0.541 & 0.590 & 0.557 & 0.595 & 0.549 & 0.590 \\
SonyAIBORobotSurface1 & 0.743 & 0.723 & 0.615 & 0.559 & 0.601 & 0.708 & 0.550 & 0.648 \\
Strawberry & 0.957 & 0.965 & 0.964 & 0.962 & 0.957 & 0.957 & 0.948 & 0.961 \\
ToeSegmentation1 & 0.792 & 0.711 & 0.778 & 0.706 & 0.715 & 0.743 & 0.735 & 0.727 \\
ToeSegmentation2 & 0.715 & 0.372 & 0.267 & 0.359 & 0.438 & 0.649 & 0.574 & 0.421 \\
Wafer & 0.995 & 0.996 & 0.996 & 0.995 & 0.988 & 0.994 & 0.967 & 0.995 \\
Wine & 0.642 & 0.685 & 0.716 & 0.568 & 0.593 & 0.543 & 0.574 & 0.586 \\
WordSynonyms & 0.583 & 0.639 & 0.608 & 0.634 & 0.579 & 0.610 & 0.601 & 0.646 \\
Yoga & 0.818 & 0.826 & 0.828 & 0.811 & 0.817 & 0.815 & 0.804 & 0.802 \\
\midrule
\textbf{Average} & 0.725 & TODO & TODO & \textbf{TODO} & TODO & TODO & TODO & \textbf{TODO} \\
\bottomrule
\end{tabular}
\end{table*}

\begin{table*}[!t]
\centering
\scriptsize
\renewcommand{\arraystretch}{0.95}
\caption{\textbf{Test accuracy for Bar and Scatter chart types under four rendering settings (image size = 128$\times$128).}}

\label{tab:bar_scatter_settings_scaled}
\begin{tabular}{l|cccc|cccc}
\toprule
\multirow{2}{*}{\textbf{Dataset}} & 
\multicolumn{4}{c|}{\textbf{Bar}} & 
\multicolumn{4}{c}{\textbf{Scatter}} \\
& Monochrome w Label & Monochrome w/o Label & Color w Label & Color w/o Label & Monochrome w Label & Monochrome w/o Label & Color w Label & Color w/o Label \\
\midrule
Adiac & 0.594 & 0.585 & 0.583 & 0.588 & 0.569 & 0.633 & 0.621 & 0.607 \\
Arrowhead & 0.731 & 0.611 & 0.718 & 0.594 & 0.690 & 0.556 & 0.691 & 0.549 \\
Beef & 0.456 & 0.456 & 0.433 & 0.544 & 0.456 & 0.533 & 0.589 & 0.556 \\
Beetlefly & 0.617 & 0.650 & 0.733 & 0.683 & 0.817 & 0.667 & 0.733 & 0.450 \\
ChlorineConcentration & 0.593 & 0.537 & 0.597 & 0.588 & 0.588 & 0.598 & 0.600 & 0.600 \\
Computers & 0.737 & 0.691 & 0.695 & 0.712 & 0.856 & 0.875 & 0.843 & 0.856 \\
CricketX & 0.601 & 0.661 & 0.616 & 0.668 & 0.632 & 0.667 & 0.638 & 0.668 \\
CricketY & 0.597 & 0.669 & 0.604 & 0.660 & 0.603 & 0.687 & 0.628 & 0.685 \\
CricketZ & 0.686 & 0.698 & 0.655 & 0.717 & 0.674 & 0.707 & 0.669 & 0.695 \\
Crop & 0.714 & 0.719 & 0.713 & 0.714 & 0.710 & 0.681 & 0.709 & 0.692 \\
ECG5000 & 0.929 & 0.913 & 0.926 & 0.933 & 0.934 & 0.931 & 0.933 & 0.932 \\
Earthquakes & 0.727 & 0.765 & 0.760 & 0.746 & 0.734 & 0.722 & 0.736 & 0.710 \\
FaceAll & 0.745 & 0.841 & 0.766 & 0.790 & 0.763 & 0.753 & 0.763 & 0.792 \\
FacesUCR & 0.845 & 0.862 & 0.833 & 0.854 & 0.836 & 0.858 & 0.839 & 0.827 \\
FordB & 0.701 & 0.763 & 0.712 & 0.750 & 0.665 & 0.705 & 0.665 & 0.699 \\
GunPoint & 0.876 & 0.927 & 0.882 & 0.873 & 0.956 & 0.953 & 0.953 & 0.962 \\
Ham & 0.625 & 0.705 & 0.654 & 0.654 & 0.606 & 0.644 & 0.629 & 0.651 \\
Herring & 0.505 & 0.531 & 0.531 & 0.531 & 0.536 & 0.552 & 0.526 & 0.542 \\
InsectWingbeatSound & 0.624 & 0.674 & 0.622 & 0.674 & 0.584 & 0.655 & 0.583 & 0.655 \\
ItalyPowerDemand & 0.949 & 0.943 & 0.958 & 0.951 & 0.928 & 0.947 & 0.942 & 0.943 \\
Lightning2 & 0.743 & 0.721 & 0.694 & 0.781 & 0.727 & 0.754 & 0.699 & 0.749 \\
PhalangesOutlinesCorrect & 0.780 & 0.786 & 0.778 & 0.794 & 0.774 & 0.764 & 0.760 & 0.756 \\
RefrigerationDevices & 0.552 & 0.612 & 0.565 & 0.586 & 0.564 & 0.550 & 0.568 & 0.560 \\
SonyAIBORobotSurface1 & 0.509 & 0.599 & 0.631 & 0.525 & 0.605 & 0.671 & 0.592 & 0.697 \\
Strawberry & 0.950 & 0.959 & 0.944 & 0.951 & 0.961 & 0.955 & 0.945 & 0.956 \\
ToeSegmentation1 & 0.703 & 0.770 & 0.740 & 0.768 & 0.735 & 0.750 & 0.662 & 0.842 \\
ToeSegmentation2 & 0.654 & 0.354 & 0.397 & 0.605 & 0.710 & 0.690 & 0.772 & 0.241 \\
Wafer & 0.987 & 0.996 & 0.996 & 0.997 & 0.994 & 0.997 & 0.997 & 0.994 \\
Wine & 0.512 & 0.562 & 0.549 & 0.574 & 0.463 & 0.525 & 0.525 & 0.519 \\
WordSynonyms & 0.613 & 0.619 & 0.594 & 0.603 & 0.620 & 0.646 & 0.632 & 0.641 \\
Yoga & 0.815 & 0.811 & 0.816 & 0.809 & 0.830 & 0.820 & 0.821 & 0.827 \\
\midrule
\textbf{Average} & TODO & TODO & TODO & \textbf{TODO} & TODO & TODO & 0.763 & \textbf{TODO} \\
\bottomrule
\end{tabular}
\end{table*}

\clearpage
\begin{table}[H]
\centering
\normalsize
\renewcommand{\arraystretch}{1.3}
\caption{\textbf{Best chart type and rendering setting for each dataset.} 
Datasets are grouped by task type and sorted by accuracy (highest to lowest within each group).}
\label{tab:best_performance_per_dataset}
\begin{tabular}{llcc}
\toprule
\textbf{Dataset} & \textbf{Best Chart} & \textbf{Best Setting} & \textbf{Accuracy (\%)} \\
\midrule
\multicolumn{4}{c}{\textbf{Binary Classification}} \\
\midrule
Wafer & Scatter & Color-L & 99.66 \\
GunPoint & Line & Color-NL & 97.11 \\
Strawberry & Line & Mono-NL & 96.49 \\
ItalyPowerDemand & Area & Color-NL & 96.05 \\
ECG5000 & Line & Mono-NL & 93.84 \\
ToeSegmentation1 & Scatter & Color-NL & 84.21 \\
Beetlefly & Area & Mono-L & 83.33 \\
Yoga & Scatter & Mono-L & 82.96 \\
Lightning2 & Bar & Color-NL & 78.14 \\
FordB & Line & Color-NL & 78.85 \\
Earthquakes & Bar & Mono-NL & 76.50 \\
PhalangesOutlinesCorrect & Line & Color-L & 80.03 \\
SonyAIBORobotSurface1 & Line & Mono-L & 74.27 \\
Wine & Line & Color-L & 71.60 \\
Ham & Bar & Mono-NL & 70.48 \\
Computers & Scatter & Mono-NL & 87.47 \\
Herring & Line & Mono-NL & 56.25 \\
ToeSegmentation2 & Scatter & Color-L & 77.18 \\
\midrule
\multicolumn{4}{c}{\textbf{Multiclass Classification}} \\
\midrule
FacesUCR & Area & Color-NL & 86.94 \\
FaceAll & Bar & Mono-NL & 84.08 \\
Arrowhead & Area & Mono-L & 73.90 \\
Crop & Area & Color-L & 72.58 \\
CricketX & Line & Mono-NL & 67.78 \\
CricketY & Line & Mono-NL & 69.40 \\
CricketZ & Line & Color-NL & 73.68 \\
ChlorineConcentration & Line & Mono-NL & 67.58 \\
WordSynonyms & Area & Color-NL & 64.58 \\
Adiac & Line & Mono-L & 64.19 \\
InsectWingbeatSound & Line & Mono-NL & 68.96 \\
Beef & Scatter & Color-L & 58.89 \\
RefrigerationDevices & Bar & Mono-NL & 61.16 \\
\bottomrule
\end{tabular}
\end{table}




\end{document}